%% file: main.tex
\documentclass[journal]{IEEEtran}
\usepackage{amsmath,amsfonts}
\usepackage[ruled, linesnumbered,lined,boxed,commentsnumbered]{algorithm2e}
\usepackage[dvipsnames,x11names,table]{xcolor}
\usepackage{svg}
\usepackage[
colorlinks,
linkcolor=blue,
citecolor=blue,
filecolor=red,
urlcolor=DodgerBlue4]{hyperref} %
\usepackage{array}
\usepackage[font=footnotesize,labelsep=period]{caption}
\usepackage{subcaption}
\usepackage{textcomp}
\usepackage{stfloats}
\usepackage{url}
\usepackage{verbatim}
\usepackage{duckuments}
\usepackage{graphicx}
\usepackage{balance}
\usepackage{cite}
\usepackage{tikz}
\usetikzlibrary{shapes.geometric} %
\usepackage{upgreek}
\usepackage{xargs}                      %

\newcommand{\xxnote}[3]{}
\newcommand{\bradynote}[1]{}

\newcommand{\nayananote}[1]{}

\ifx\hidenotes\undefined
  \usepackage{color}
  \renewcommand{\xxnote}[3]{\color{#2}{#1: #3}}
\fi

\ifx\hidenotes\undefined
    \renewcommand{\bradynote}[1]{\todo{BM: #1}}
    \renewcommand{\nayananote}[1]{\info{Nay: #1}}
\fi

\definecolor{tabfirst}{rgb}{1, 0.7, 0.7} %
\definecolor{tabsecond}{rgb}{1, 0.85, 0.7} %
\definecolor{tabthird}{rgb}{1, 1, 0.7} %

\newcommand{\PlannerName}{\mbox{IA-TIGRIS}}
\newcommand{\PlannerNameSpaced}{\mbox{IA-TIGRIS }}

\begin{document}

\title{\PlannerName: An Incremental and Adaptive Sampling-Based Planner for Online \\Informative Path Planning}

\author{Brady Moon, Nayana Suvarna, Andrew Jong, Satrajit Chatterjee, Junbin Yuan, \\Muqing Cao, and Sebastian Scherer
\thanks{Received November 30, 2025; accepted February 4, 2026. This work is supported by the Office of Naval Research (Grant N00014-21-1-2110) and the National Science Foundation Graduate Research Fellowship under Grant No. DGE1745016. This article was recommended for publication by Associate Editor C. Verginis and Editor P. Robuffo Giordano upon evaluation of the reviewers' comments.  \textit{(Corresponding author: Brady Moon)}}%
\thanks{Brady Moon was with the Robotics Institute, School of Computer Science at Carnegie Mellon University, Pittsburgh, PA 15213 USA. He is now with the Department of Mechanical Engineering, Brigham Young University, Provo, UT 84602 USA (email: brady.moon@byu.edu).}%
\thanks{Nayana Suvarna, Andrew Jong, Muqing Cao, and Sebastion Scherer are with the Robotics Institute, School of Computer Science, Carnegie Mellon University, Pittsburgh, PA 15213 USA.}
\thanks{Satrajit Chatterjee was with the GRASP Lab at the University of Pennsylvania, Philadelphia, PA 19204 USA.}%
\thanks{Jungin Yuan is with the Mechanical Engineering Department at Carnegie Mellon University, Pittsburgh, PA 15213 USA.}%
\thanks{Digital Object Identifier 10.1109/TRO.2026.3672542}
}

\IEEEpubid{\begin{minipage}{\textwidth}\ \\[12pt]
\centering
  \copyright~2026 IEEE Personal use of this material is permitted.  Permission from IEEE must be obtained for all other uses, in any current or future media, including reprinting/republishing this material for advertising or promotional purposes, creating new collective works, for resale or redistribution to servers or lists, or reuse of any copyrighted component of this work in other works.
\end{minipage}} 

\maketitle

\begin{abstract}

Planning paths that maximize information gain for robotic platforms has wide-ranging applications and significant potential impact. 
To effectively adapt to real-time data collection, informative path planning must be computed online and be responsive to new observations. 
In this work, we present \PlannerNameSpaced (Incremental and Adaptive Tree-based Information Gathering Using Informed Sampling), which is an incremental and adaptive sampling-based informative path planner designed for real-time onboard execution. Our approach leverages past planning efforts through incremental refinement while continuously adapting to updated belief maps. We additionally present detailed implementation and optimization insights to facilitate real-world deployment, along with an array of reward functions tailored to specific missions and behaviors.
Extensive simulation results demonstrate \PlannerNameSpaced generates higher-quality paths compared to baseline methods.
We validate our planner on two distinct hardware platforms: a hexarotor unmanned aerial vehicle (UAV) and a fixed-wing UAV, each having different motion models and configuration spaces.
Our results show up to a 38\% improvement in information gain compared to baseline methods, highlighting the planner's potential for deployment in real-world applications. 
Project website: \href{https://ia-tigris.github.io}{ia-tigris.github.io}.

\end{abstract}


\input{tex/1_intro}

\input{tex/2_problem}

\input{tex/3_methodology}

\input{tex/4_evaluation}

\input{tex/5_results}

\input{tex/6_conclusion}

\bibliography{references}

\begin{IEEEbiography}[{\includegraphics[width=1in,height=1.25in,clip,keepaspectratio]{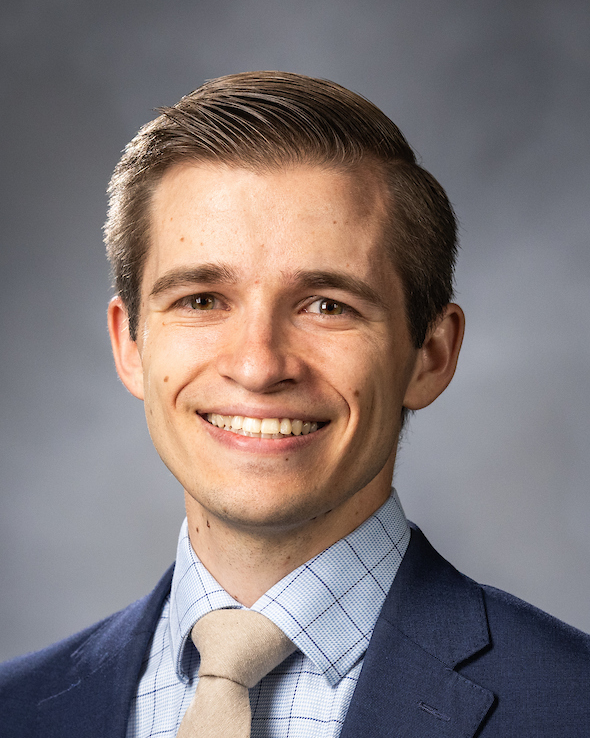}}]{Brady Moon}
is an Assistant Professor at Brigham Young University. 
His research focuses on autonomous information gathering and decision-making for field robotics, spanning task and motion planning, multi-agent coordination, human-robot interaction, and learning-enabled autonomy. 
He received his B.S. in Electrical Engineering from Brigham Young University in 2019 and his Ph.D. in Robotics from Carnegie Mellon University in 2025.
\end{IEEEbiography}
\vspace{-33pt}
\begin{IEEEbiography}[{\includegraphics[width=1in,height=1.25in,clip,keepaspectratio]{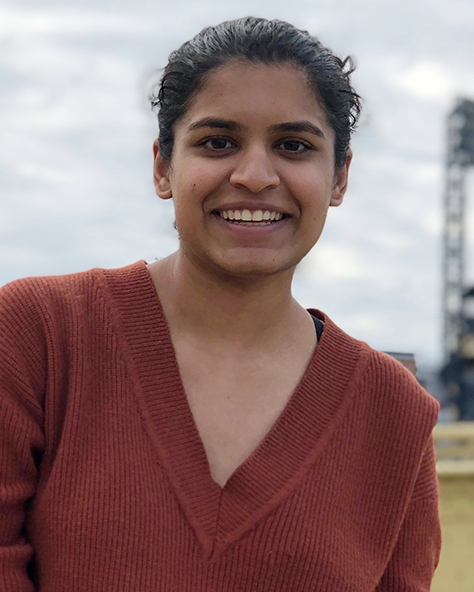}}]{Nayana Suvarna}
received her Masters in Robotics at Carnegie Mellon University in 2025. Her research focused on information gathering algorithms for teams of aerial vehicles. Nayana holds a B.S. in Computer Engineering from the University of Pittsburgh.
\end{IEEEbiography}
\vspace{-33pt}
\begin{IEEEbiography}[{\includegraphics[width=1in,height=1.25in,clip,keepaspectratio]{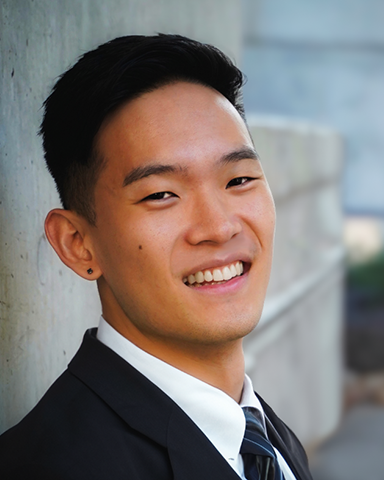}}]{Andrew Jong}
is a Ph.D. student in Robotics at Carnegie Mellon University. His research focuses on robotics for disaster response, and spans planning, perception in degraded environments, agile navigation, and safety. Andrew holds a B.S. in Computer Science from San Jos\'e State University and a M.S. in Robotics from Carnegie Mellon University.
\end{IEEEbiography}
\vspace{-33pt}
\begin{IEEEbiography}[{\includegraphics[width=1in,height=1.25in,clip,keepaspectratio]{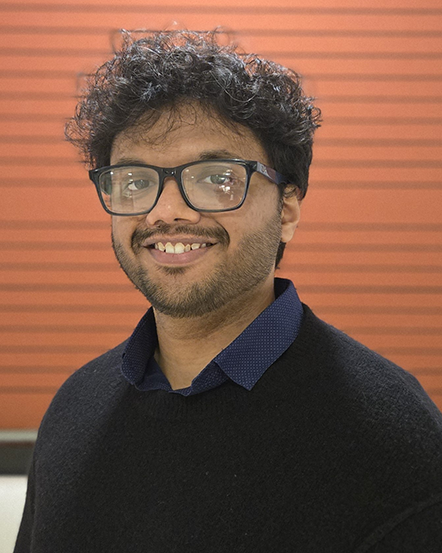}}]{Satrajit Chatterjee}
has a Master's in Robotics from the GRASP Lab, University of Pennsylvania. His research focused on path-planning for robotics and includes work in multi-target search and tracking, and vision-based generative techniques for path-planning in semantically complex environments. Satrajit holds a B.Tech in Computer Science and Engineering from the National Institute of Technology, Tiruchirappalli, India. 
\end{IEEEbiography}
\vspace{-33pt}
\begin{IEEEbiography}[{\includegraphics[width=1in,height=1.25in,clip,keepaspectratio]{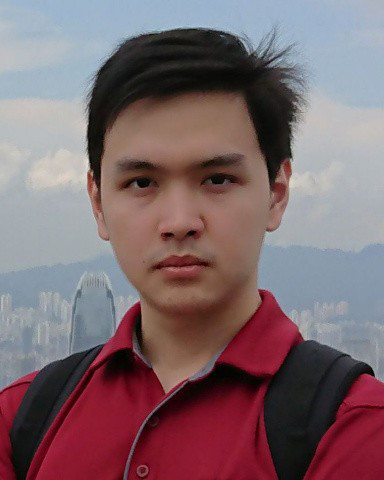}}]{Junbin Yuan}
is a Ph.D. candidate in mechanical engineering at Carnegie Mellon University. His research focuses on planning algorithms for robotics, and covers coverage path planning and multi-target search and tracking. Junbin received his B.Eng. in Electronics Engineering from Hong Kong University of Science and Technology 
\end{IEEEbiography}
\vspace{-33pt}
\begin{IEEEbiography}[{\includegraphics[width=1in,height=1.25in,clip,keepaspectratio]{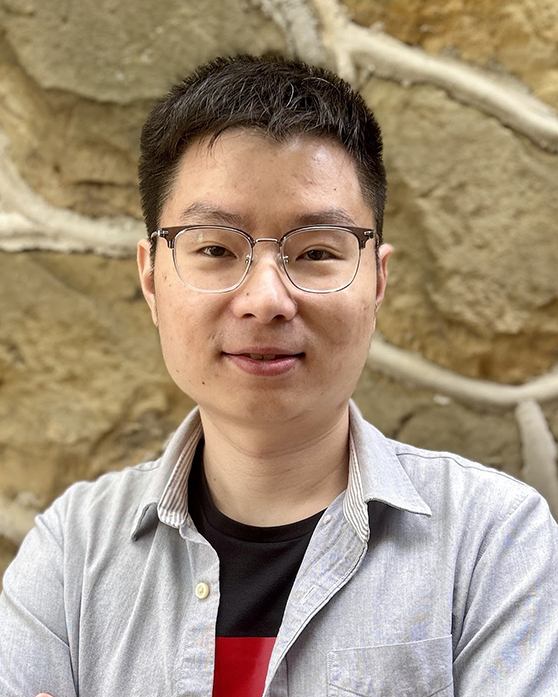}}]{Muqing Cao}
is a Postdoctoral Fellow at the Robotics Institute, Carnegie Mellon University. He received his Ph.D. in Electrical and Electronic Engineering from Nanyang Technological University (NTU), Singapore. His research focuses on multi-robot systems, task and motion planning, and field robotics. He has received the Robotics: Science and Systems (RSS) Pioneer 2025 and the IROS 2023 Best Entertainment and Amusement Paper Award.
\end{IEEEbiography}
\vspace{-33pt}
\begin{IEEEbiography}[{\includegraphics[width=1in,height=1.25in,clip,keepaspectratio]{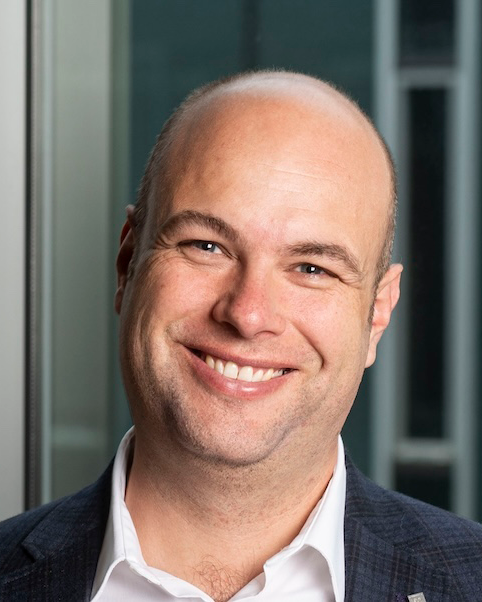}}]{Sebastian Scherer} 
 is an Associate Research Professor at the Robotics Institute (RI) at Carnegie Mellon University (CMU). His research focuses on enabling autonomy in challenging environments and previously led CMU’s entry to the SubT challenge. He and his team have shown several firsts for autonomy for flying robots and off-road driving . Dr. Scherer received his B.S. in Computer Science, M.S. and Ph.D. in Robotics from CMU in 2004, 2007, and 2010. 
\end{IEEEbiography}

\vfill

\end{document}

%% file: tex/1_intro.tex
\section{Introduction}

Robots play a vital role in gathering information from the physical world, supporting a wide range of applications such as scientific research \cite{doi:10.1126/scirobotics.abc3000,McCammon2021,de2021monitoring}, environmental monitoring \cite{kaufmann2021conventional, 10161136,barbedo2019review, Christensen2015,patrikar2020}, search and rescue operations \cite{Bashyam2019UAVsFW,Alsamhi2022}, and disaster response efforts \cite{bejiga2017convolutional,Mohsan2023,BAILONRUIZ2022104071}. By employing intelligent algorithms, robotic systems enhance the efficiency of data collection, provide valuable insights, and support well-informed decision-making processes. These autonomous robots provide unparalleled advantages in situations where human access is constrained, dangerous, or logistically challenging \cite{6161683}. Moreover, the scalability of autonomous information-gathering robots enhances the rate of information gathering by not capping the number of robots based on the number of human operators. 

\begin{figure}[t]
\centering
\includegraphics[width=\columnwidth]{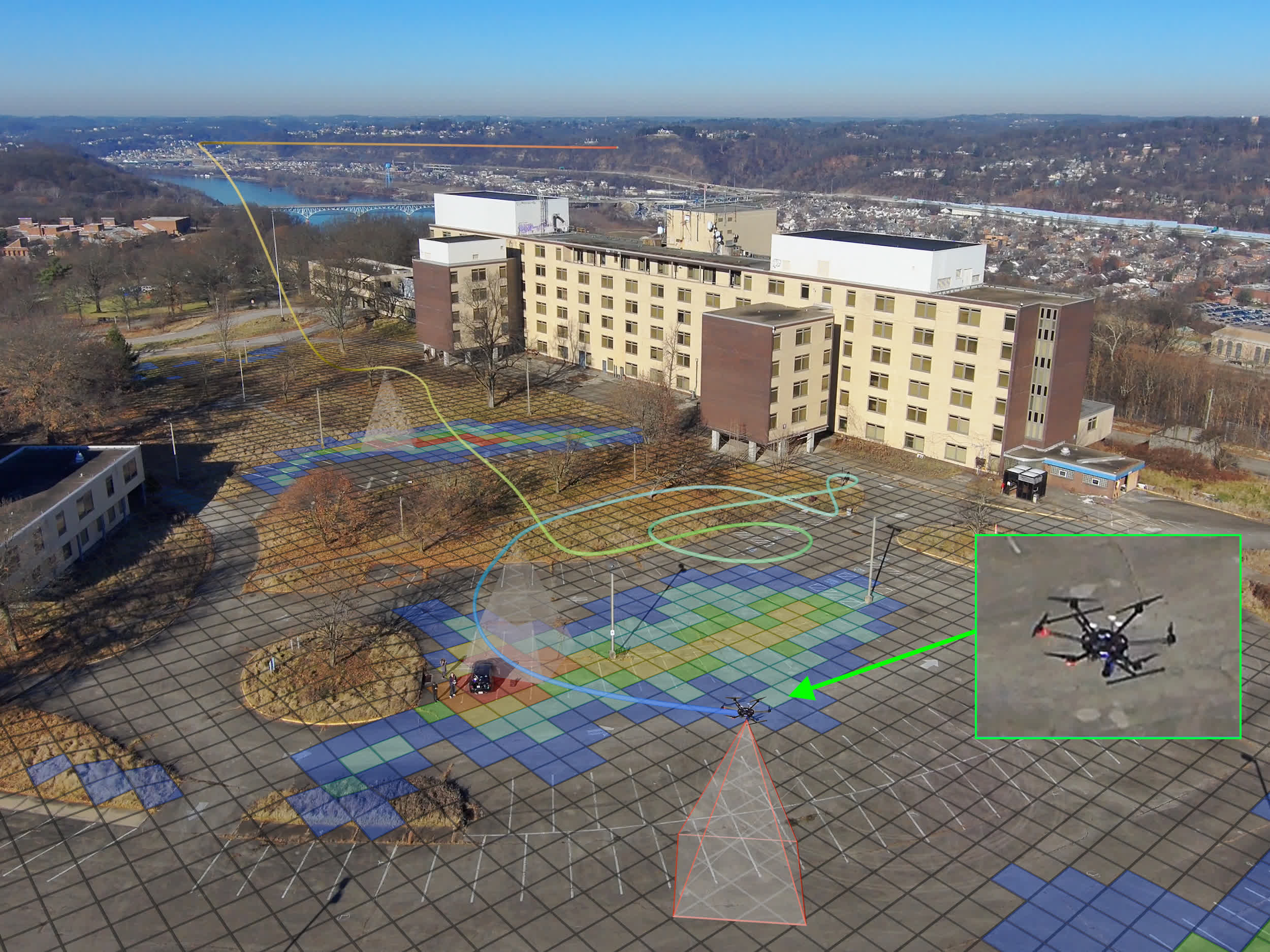}
\caption{\PlannerNameSpaced deployed on a hexarotor UAV to map the location of cars in the environment. The visualization shows a probability grid representing the prior belief of car locations and a representative path generated by our planner. The algorithm runs entirely onboard, continuously refining and adapting paths based on updated world beliefs from sensor observations.}
\label{fig:fig1}
\end{figure}

\IEEEpubidadjcol
Informative path planning (IPP) seeks to create intelligent paths for robots that maximize their inherent potential and advantages in data-gathering situations, while not violating budget constraints on the system or mission. This approach has many advantages over most coverage-based approaches, which often do not take into account the flight time of the drone or prior information about the space and do not adapt to information learned during flight. However, solving the IPP problem is computationally expensive, being at least NP-hard even in a discrete space \cite{singh2009}, leading to methods that plan only over a short horizon or are static during the path execution.

Recent works in IPP have explored learning-based approaches \cite{cao2023catnipp, harutyunyan2025mapexrl} that target longer planning horizons and graph-based methods that optimize routes over fixed spatial discretizations \cite{OTT2024104814, Dutta2025informative}. While effective, many still rely on fixed-resolution search graphs that can be difficult to scale to large, high-dimensional spaces. They also often use simplified sensor models (e.g., pointwise measurements or circular downward-facing fields of view), assume independent observation rewards without accounting for diminishing returns from repeated observations, and may ignore information gathered along continuous trajectories. In addition, learning-based methods can struggle when environmental conditions differ from those encountered during training \cite{POPOVIC2024104727}.

To address these limitations, \cite{moon2022tigris} presented an informed, sampling-based algorithm that optimizes over the current belief map without requiring prior training. The approach scales to high-dimensional spaces, supports realistic sensor configurations, reasons over edge rewards to capture information gathered along the entire trajectory, and accounts for reward dependence, including multiple observations of the same state. However, plans were static and had to be recomputed from scratch whenever the environment changed.

In this work, we introduce \PlannerNameSpaced (Incremental and Adaptive Tree-based Information Gathering Using Informed Sampling), which is an adaptive and incremental sampling-based informative path planner. \PlannerNameSpaced builds on and substantially improves the approach in \cite{moon2022tigris} by introducing new tree-building logic and novel belief map node embeddings, enabling efficient incremental planning as new information is gathered.
\PlannerNameSpaced operates fully online, efficiently updating and building its set of possible trajectories and directly incorporating changes in the belief map. This adaptive, incremental approach leads to global paths that are more robust to path execution disturbances, better adapt to evolving environment knowledge, and achieve higher quality than a single static plan through continual online refinement. Fig.~\ref{fig:fig1} demonstrates \PlannerNameSpaced deployed on a hexarotor drone for an IPP application.

Our key contributions are the following:

\begin{itemize}
    \item A novel incremental and adaptive sampling-based planner for IPP that dynamically plans online using the updated belief map. We provide key implementation details, insights, and optimizations that enable efficient trajectory evaluation and updates during incremental planning.
    \item Thorough testing and evaluation of our method against other baselines, as well as ablations and analysis on the components of our algorithm.
    \item Hardware validation on two distinct UAV platforms with different motion constraints and configuration spaces, demonstrating versatility and real-world applicability.
    \item Release of our open-source codebase, simulation environment, and testing package for community use, development, benchmarking, and testing.
\end{itemize}

The paper is organized as follows: Section \ref{sec:related} gives a detailed overview of related works and the contributions of this work. Section \ref{sec:problem} presents the problem formulation. Our proposed approach is detailed in Section \ref{sec:approach}. That section outlines the planner's design and its integration with system components to enable adaptive and efficient IPP for real-world applications. Section \ref{sec:results} provides our detailed experimental evaluation and results of the approach. The results of our field deployments are summarized in Section \ref{sec:field}, and Section \ref{sec:conclusion} offers the conclusion and future work.

\section{Related Works}\label{sec:related}

One way to formulate the IPP problem is to discretize the continuous space into a fixed set of states. Many previous works have then used this discrete space to view IPP as an orienteering problem. Given a weighted-undirected graph, orienteering involves determining a Hamiltonian path over a subset of nodes 
that maximizes total reward while satisfying a path cost constraint. The orienteering problem combines the objective of selecting a subset of nodes from the graph and solving a traveling salesman problem to minimize distance while visiting the selected nodes. The optimization objective for node selection can be tailored to specific applications. For example, \cite{lorenzo_orienteering} applied it to environmental monitoring in an aquatic setting while \cite{arora_randomized_2017} applied it to the more general constraint satisfaction problem. 

However, framing IPP as an orienteering problem leads to computationally expensive runtimes, which rapidly become intractable as the likelihood of near-optimal paths in the solution space diminishes. As \cite{efficient-sensing} noted, the IPP problem is a NP-hard search problem. To overcome this problem complexity, they developed an approximation algorithm that efficiently finds near-optimal solutions by leveraging mutual information in the problem space. They ensured that the mutual information formulation is submodular in nature \cite{submodular-krause}. This submodular nature creates a diminishing returns property where each additional observation provides less new information than the previous one. Hence, the IPP problem can be reformulated as maximizing a submodular objective function while constrained to a budget. \cite{recursive_greedy} studied this formulation by developing a recursive-greedy algorithm with strong theoretical guarantees.

However, these approaches have been used only in applications that have relatively small search spaces, such as environmental monitoring \cite{lorenzo_orienteering}, water quality analysis \cite{water-1}, or chemical plume detection \cite{1703649} that are discretized into a finite set of sensing locations. These approaches quickly become intractable in larger, higher-dimensional applications, such as searching for targets over large areas (e.g., missing hikers) or tracking multiple dynamic targets (e.g., ships at sea).  To address this intractability, \cite{lorenzo_orienteering} proposed using heuristics, while \cite{lorenzo_orienteering, singh2009, recursive_greedy} suggested using greedy approaches to approximate solutions. Mixed-integer programming solutions \cite{mip_forumation} have also been proposed. But these solutions fail to capture the relationship between nodes, which would be problematic for many applications. The limitations in capturing rewards relationships among nodes, combined with inefficient runtimes, necessitate developing better exploration algorithms for robust information gathering by autonomous systems in complex settings. 

Some previous approaches have used receding-horizon-based methods to solve the IPP problem. Local receding-horizon solutions optimize the information objective over a small lookahead horizon, relying on replanning to approach global optimality. One approach involves a receding-horizon algorithm \cite{receding} that satisfies temporal logic specifications for safety-critical applications. Another approach \cite{no-regret} uses a receding-horizon planner for informative planning in a latent environment modeled using Gaussian processes. According to \cite{experimental_receeding_horizon}, receding-horizon-based path planners are susceptible to becoming trapped in local optima. To address this, they explored adjusting horizon length and propose a J-Horizon algorithm that incorporates lookahead step size to improve convergence toward better optima. Despite these improvements, receding-horizon methods primarily plan optimal paths reactively and fail to fully utilize the available prior data \cite{Kailas-2023-137637}.

Notably, \cite{frolov2014} compared lawnmower paths with other planning algorithms and found that lawnmower paths perform nearly as well as adaptive algorithms and outperform graph-based search algorithms that struggle to adapt to uncertainties in prior information. The ability to allow for revisits is very important for finding globally optimal paths, especially when the information objective is submodular or when the task requires revisits \cite{moon2022tigris}. Branch and bound techniques show promise by pruning suboptimal branches early in the tree search \cite{6224902, bnb-2}, but efficiently calculating tight bounds in unknown, high-dimensional environments remains challenging. While Gaussian processes have been used to address this, implementing Gaussian processes is application specific and the computational complexity scales cubically with the number of observations \cite{rasmussen2006gaussian}. Although methods exist to mitigate this complexity, long-horizon planning remains computationally demanding unless stationary assumptions are made about the covariance function. This complexity is further compounded when planning in higher-dimensional state spaces.

Recent studies have explored learning-based and optimization-based approaches for enabling long-horizon IPP. Learning-based methods leverage offline-trained neural networks or policies to guide online decision-making. Specifically, \cite{Rückin2022adaptive} enhanced online tree search by utilizing a network trained offline to efficiently sample informative robot actions. \cite{Choi2021adaptive} developed a learning-based controller that dynamically switches among multiple classical local search strategies to enhance robustness in various scenarios. \cite{cao2023catnipp} presented a context-aware attention-based network for informative path planning (CAtNIPP), which is trained to select subsequent robot states within a predefined sensing graph, while \cite{Vashisth2024deep} extended this concept to dynamically select actions within an online-generated sensing graph.
Optimization-based approaches explicitly model the information gain as an optimization objective and utilize numerical optimization techniques for planning. Examples include mixed-integer programming \cite{Dutta2025informative}, Monte Carlo tree search (MCTS) \cite{Ott2023sequential}, and dynamic programming \cite{OTT2024104814}. Notably, the dynamic programming approach in \cite{OTT2024104814} achieves efficiency through approximation strategies, such as assuming the robot teleports directly to future positions and calculating information gain based solely on current map entropy.
However, most existing methods assume simplified point-based sensor models, significantly differing from realistic scenarios where sensor footprints are nontrivial, application-specific, and governed by practical sensor parameters. Moreover, these approaches generally overlook the robot's heading and kinematic constraints, which substantially influence real-world planning outcomes.

Numerous works address these drawbacks by formulating the task of information gathering in continuous space and introducing sampling-based algorithms. Examples of some of these works include \cite{hollinger_sampling-based_2014}, \cite{schmid_efficient_2020}, \cite{hollinger_long-horizon_2015, nasa-mcts, adaptive-sampling-1, adaptive-sampling-2, 6907763, 6630605}. These methods employ sampling-based approaches, which involve the selection and inclusion of new states in continuous space, whether visiting all the sampled states or adding them into a tree structure of potential trajectories. As the tree grows, the optimal path---maximizing information gain while respecting budget constraints---is continuously refined and the best path is selected. However, these approaches can face significant challenges in expansive spaces, where increasing dimensionality causes the search space to grow substantially relative to regions of high information reward. In order to address these challenges, \cite{moon2022tigris} introduced a sampling-based approach for IPP in large and high-dimensional search spaces. The method performs informed sampling within the continuous space and incorporates graph-edge information gain during reward estimation, efficiently generating global paths that optimize information gathering within budget constraints.

Another crucial aspect of IPP planners is their broad categorization into adaptive and nonadaptive planners. Following prior works such as \cite{popovic2020informative, POPOVIC2024104727,hollinger_long-horizon_2015,doi:10.1177/0278364912467485,https://doi.org/10.1002/rob.21722, Vashisth2024deep}, we define adaptive planning methods as those that do not follow a static, predefined path but instead continue to modify the plan during execution. This adaptivity enables the planner to respond to discrepancies between expected and observed measurements, to adjust for execution-time disturbances, and to refine the plan online as more information becomes available. Adaptivity is often implemented through frequent replanning and belief updates. 
While this approach can improve performance, it increases computational complexity and requires lightweight planners capable of real-time execution on robotic platforms

Our proposed algorithm, IA-TIGRIS, is an adaptive planner that builds upon the work of \cite{moon2022tigris}, which had demonstrated effective global planning but generated static plans that remained unchanged during robot execution. IA-TIGRIS also uses a sampling-based framework but continuously optimizes and refines plans online as the robot acquires new information. This adaptive refinement is crucial for long-horizon planning scenarios where optimal paths are difficult to determine in a single planning cycle. In order to achieve sufficient computational efficiency for onboard deployment, we introduce improvements in the tree-building logic and add novel belief map node embeddings that facilitate efficient incremental planning. We validate our method through extensive simulation testing and demonstrate efficient online planning capabilities on hardware across two distinct UAV systems.

%% file: tex/2_problem.tex
\section{Problem Formulation}\label{sec:problem}

This work addresses the challenge of maximizing data gathered by a robot within the specified budget constraints of the system, a problem formally known as the IPP problem. 
Following \cite{moon2022tigris}, we define it as
\begin{equation*}
    \mathcal{T}^*=\underset{\mathcal{T}\in \mathbb{T}}{\arg\max}~I(\mathcal{T}) \text{ s.t.\ $C(\mathcal{T}) \leq B$}.
\end{equation*}

Here, $\mathcal{T}$ represent the sensor trajectory from the set of feasible trajectories $\mathbb{T}$, and $C(\mathcal{T})$ is its cost. Examples of potential cost functions include trajectory length, energy consumption, and flight time, and each would not exceed the budget constraint $B$. Lastly, $I(\mathcal{T})$ denotes the reward or information gain from trajectory $\mathcal{T}$, and the goal is to find the optimal trajectory $\mathcal{T}^*$ that maximizes information gain without exceeding the budget. 

In this work, trajectories are planned in a four-dimensional space defined by $(x, y, z, \psi)$, where $x, y, z$ specify the robot’s position and $\psi$ its heading. We typically sample new states from a discrete altitude or set of altitudes (for $z$), while $x, y, \psi$ remain continuous. In our experiments, the cost $C(\mathcal{T})$ is the total distance traveled by the robot. The information gain function $I(\mathcal{T})$ computes expected reduction in uncertainty over the environment, modeled as a belief map representing the probability of object presence across spatial cells. While in our applications this belief map encodes discrete objects of interest, the algorithm itself is agnostic to the underlying representation. Other implementations could use different belief models, environment features, or custom metrics for information gain. Importantly, our reward function is not constrained to be modular or submodular, allowing for more expressive and task-specific definitions of information gain.

%% file: tex/3_methodology.tex
\section{Proposed Approach}\label{sec:approach}

\begin{figure}[t]
\centering
\includegraphics[trim={0cm 0cm 0cm 0cm},clip,width=\columnwidth]{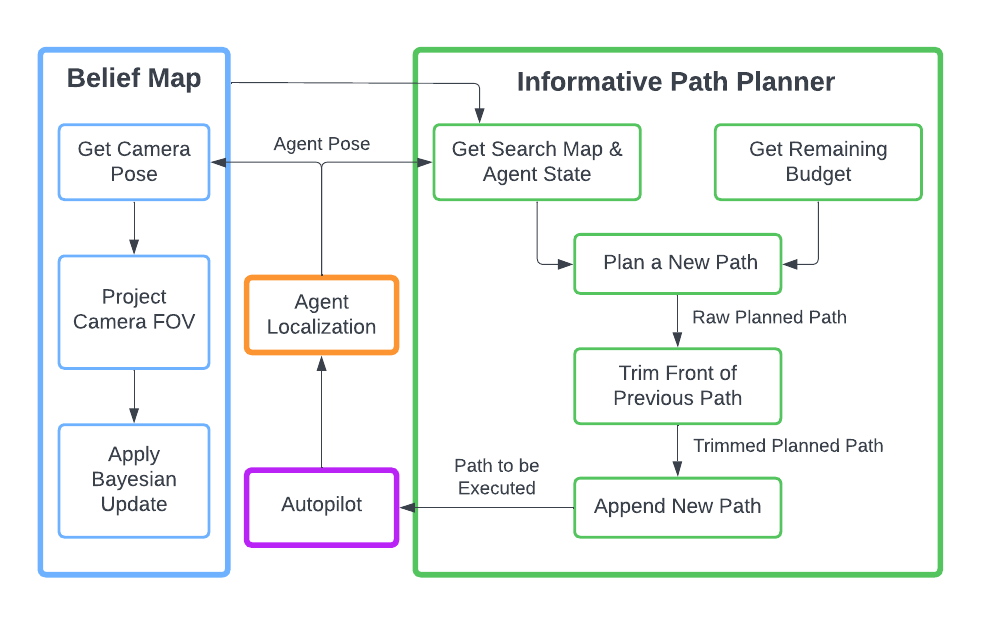}
\caption{
The IA-TIGRIS planning framework consists of two main components: the belief map and the informative path planner. The belief map is responsible for managing the agent's understanding of the environment while the informative path planner generates adaptive plans based on the agent's position and remaining budget. For the planning procedure, the new path is merged with the existing plan by pruning outdated portions of the existing path and starting the new plan from the agent's expected position.
}
\label{fig:planner_diagram}
\end{figure}

This section details our proposed approach for IPP online using incremental and adaptive methods. Section \ref{sec:alg_overview} provides a thorough explanation of the core algorithm, including the main planning loop and the graph update procedure, elucidating the steps involved in sampling, steering, and constructing the tree of potential paths. Section \ref{sec:replanning} discusses the strategy to enable adaptive replanning with the presented algorithm. Finally, Section \ref{sec:incr-adap-plan} explores the techniques used to incrementally update the path, enhancing the efficiency and responsiveness of the planning process. Sections \ref{sec:beliefspace} and \ref{sec:sensor-model} explain how the belief map and observation model integrate into the planner, and Section \ref{sec:reward-models} presents our reward models. A high-level overview of this framework is shown in Fig.~\ref{fig:planner_diagram}.

\begin{algorithm}[t]
\SetInd{0.4em}{0.8em}
\DontPrintSemicolon
\small

\SetKwInOut{Input}{Input}\SetKwInOut{Output}{Output}
\SetKwFunction{InformedSample}{InformedSample}

\Input{$\mathbf{x}_{start}$: Agent start pose \\
       $\mathcal{X}$: State space \\
       $B$: Budget \\
       $T$: Allowed planning time \\
       $\Delta$: Extend distance for tree building\\
       $R$: Radius for neighbor selection}

\Output{$\mathcal{T}$: Planned path}
\If{$V=\emptyset$}
{
    $I_s \leftarrow I(\mathbf{x}_{start})$; $C \leftarrow 0$\;
    $n_{start} \leftarrow \{\mathbf{x}_{start}, I_s, C\}$\;
    $V \leftarrow {n_{start}}$; $E \leftarrow \emptyset$; $V_{closed} \leftarrow \emptyset$; $G \leftarrow (V,E)$\;
}
\Else{
    \textsc{UpdateGraph}$(\mathbf{x}_{start},B)$
}

\While{computation time $< T$}
{
    $\mathbf{x}_{sample} \leftarrow$ \textsc{InformedSample}$(\mathcal{X})$\;
    $n_{nearest} \leftarrow $\textsc{Nearest}$(\mathbf{x}_{sample})$\;
    $\mathbf{x}_{feas}, e_{feas} \leftarrow $\textsc{Steer}$(n_{nearest}, \mathbf{x}_{sample}, \Delta, X_{free})$\;
    \textsc{AddPoseToGraph}$(\mathbf{x}_{feas}, e_{feas}, n_{nearest}, \Delta, R, B)$\;
    $N_{near} \leftarrow $\textsc{Near}$(\mathbf{x}_{feas}, R, V \setminus V_{closed})$\;
    \For{$n_{near} \in N_{near}$}
    {
        \If{$n_{near} \ne x_{feas}$ and $n_{near} \ne n_{nearest}$}
        {
        
            $\mathbf{x}_{new},e_{new}\leftarrow$\textsc{Steer}$(n_{near},\mathbf{x}_{feas},\!\Delta, X_{free})$\;
            \If{$\mathbf{x}_{new} \neq n_{near}$}
            {
                \textsc{AddPoseToGraph}$(\mathbf{x}_{new}, e_{new}, n_{near}, \Delta, R, B$)
            }
        }
    }
}
\Return $\mathcal{T} \leftarrow $\textsc{BestPath}$(G)$
\caption{\textsc{IA-TIGRIS} }
\label{alg:ours1}
\end{algorithm}
\begin{algorithm}[t]
\SetInd{0.4em}{0.8em}
\DontPrintSemicolon
\small

\SetKwInOut{Input}{Input}\SetKwInOut{Output}{Output}
\SetKwFunction{InformedSample}{InformedSample}

\Input{$\mathbf{x}_{new}$: New pose \\
       $e_{new}$: Edge from $n_{parent}$ to $\mathbf{x}_{new}$ \\
       $n_{parent}$: Parent node in graph \\
       $\Delta$: Extend distance for tree building \\
       $R$: Radius for neighbor selection \\
       $B$: Budget}

$I_{new} \leftarrow I(\mathbf{x}_{new}, e_{new}, n_{parent})$\;
$C_{new} \leftarrow C_{n_{parent}} + Cost(e_{new})$\;
$n_{new} \leftarrow \{\mathbf{x}_{new}, I_{new}, C_{new}\}$\;
\If{not \textsc{Prune}$(n_{new})$}
{
    $E \xleftarrow{+} \{ e_{new}\}$\;
    $V \xleftarrow{+} \{ n_{new}\}$\;
    $G \leftarrow (V,E)$\;
    \If{$C_{new} = B$}
    {
        $V_{closed} \xleftarrow{+} \{ n_{new}\}$\;
    }
}
\caption{\textsc{AddPoseToGraph}}
\label{alg:ours2}
\end{algorithm}

\subsection{Algorithm Overview}\label{sec:alg_overview}
The IA-TIGRIS algorithm comprises two main components: the main planning loop (Algorithm \ref{alg:ours1}) and the graph update procedure (Algorithm \ref{alg:ours2}). Let the notation $X \xleftarrow{+} \{\mathbf{x}\}$ and $X \xleftarrow{-} \{\mathbf{x}\}$ represent the operations $X \leftarrow X \cup \{\mathbf{x}\}$ and $X \leftarrow X \setminus \{\mathbf{x}\}$ respectively. The \PlannerNameSpaced planning loop takes as input the agent start pose, $x_{start}$, the state space $\mathcal{X}$, the remaining budget $B$, the remaining allowed planning time $T$, the extend distance for tree building $\Delta$, and the radius for neighbor selection $R$.

At the beginning of planning in Algorithm \ref{alg:ours1}, the set of vertices $V$ is empty. The initial information $I_s$ at the start pose $x_{start}$ is calculated by $I(\cdot)$, and the initial cost $C$ is set to 0 (Line 2). The tuple $n_{start}$ is initialized and added to the set of vertices $V$, while the set of edges $E$ and the closed set of vertices $V_{closed}$ are initialized as empty (Lines 3--4). The graph G is initialized as $G \leftarrow (V, E)$.

The algorithm then loops until the computation time exceeds the allowed planning time $T$ (Line 9). Each iteration draws a sample $x_{sample}$ from the state space $\mathcal{X}$ using \mbox{\textsc{InformedSample}}, which is a weighted sampler based on the reward for viewing a state in the belief map. For the grid-based belief model used in this work, this is done by first conducting the weighted sample in the belief map and then uniformly sampling in the subset of the configuration space that views the sampled belief map state while respecting desired observation constraints (see \cite{moon2022tigris} for related details). Informed sampling focuses paths on areas of higher reward rather than uniformly sampling in the search space.

The algorithm identifies the nearest node $n_{nearest}$ in the graph $G$ to $x_{sample}$ (Line 11) and attempts to steer from $n_{nearest}$ toward $x_{sample}$ within distance $\Delta$ while checking for collisions, resulting in a feasible pose $x_{feas}$ and an edge $e_{feas}$ (Line 12). Unlike \cite{moon2022tigris}, the new pose and edge are immediately evaluated for addition to the graph $G$ using \mbox{\textsc{AddPoseToGraph}} (Line 13), allowing the neighbor selection radius $R$ to be less than the extend distance $\Delta$.

The algorithm then searches for nearby nodes $N_{near}$ within a radius $R$ of $x_{feas}$ that are not in the closed set $V_{closed}$ (Line 14). For each neighbor $n_{near}$ in $N_{near}$, if $n_{near}$ is distinct from $x_{feas}$ and $n_{nearest}$, the algorithm attempts to steer from $n_{near}$ to $x_{feas}$ (Lines 15--17). If the new pose $x_{new}$ is distinct from $n_{near}$, the pose and edge are evaluated to be added to the graph $G$ using the \textsc{AddPoseToGraph} function (Lines 18--19). This neighborhood expansion structure is different from \cite{moon2022tigris} in that new expansions now allows for neighboring nodes to extend toward $x_{feas}$ even if it did not add an edge connecting to $n_{nearest}$ due to a collision or budget constraint.

After the computation time is exhausted, the algorithm extracts the best path $T$ from the graph $G$ using the \textsc{BestPath} function (Line 24). This function simply returns the path with the highest information reward in the graph. 

\begin{figure}[t]
\centering
\includegraphics[trim={0cm 0cm 0cm 0cm},clip,width=\columnwidth]{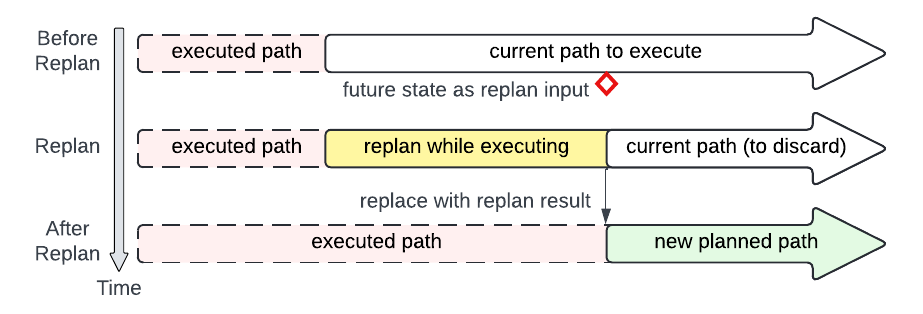}
\caption{Visualization of the online replanning procedure. The future robot state is used as the starting point for the next plan. The newly-planned path is merged with the previous path.}
\label{fig:planner_replan}
\end{figure}

The \textsc{AddPoseToGraph} function, outlined in Algorithm~\ref{alg:ours2}, handles the addition of a new pose $x_{new}$ and edge $e_{new}$ to the graph $G$. The information $I_{new}$ for the new pose is calculated, and the cumulative cost $C_{new}$ is computed by adding the cost of the edge $e_{new}$ to the cost of $n_{parent}$ (Lines 1--2). A new node $n_{new}$ is created as a tuple containing the new pose, information, and cumulative cost (Line 3). The node $n_{new}$ is then checked by the \textsc{Prune} function (Line 4). The pruning function checks nearby nodes within a radius to see whether they contain a better solution. Because most of the nodes in the tree are not full paths, the upper bound on their rewards are compared. For more aggressive pruning, heuristics can be used for this comparison. In our case, we use the heuristic of checking whether a nearby node has a lower cost and higher current reward, but this heuristic can have implications on finding the optimal solution. If $n_{new}$ passes the pruning check, the edge $e_{new}$ and node $n_{new}$ are added to the sets $E$ and $V$, respectively, and graph $G$ is updated (Lines 4--7). If the cumulative cost $C_{new}$ equals the budget $B$, the new node $n_{new}$ is added to the closed set $V_{closed}$ (Lines 8--9).

\subsection{Adaptive Planning}\label{sec:replanning}

For environments where the world is static and measurements are deterministic, global plans could be static and created once before flight. Having a single, static plan would also work well in situations when onboard computation is restricted, there is high confidence the global path is optimal, the executed path will be perfectly followed, and the predicted information gain will be close to the actual information gain. However, there are many cases where the plan would need to be adjusted due to disturbances and modified due to new information. When disturbances cause deviations from an expected observation, the path can be replanned during path execution to reobserve locations and adjust the path according to the current belief map. 

The high adaptability of \PlannerNameSpaced is due to its ability to continuously replan the global trajectory online. The updated paths take into account the current world belief updated by all observations. This adaptability brings robustness, as disturbances may cause deviations from the planned path and lead to differences between planned and expected observations. 

\begin{algorithm}[t]
\SetInd{0.4em}{0.8em}
\DontPrintSemicolon
\small

\SetKwInOut{Input}{Input}
\SetKwFunction{FindMatchingNode}{FindMatchingNode}
\SetKwFunction{PruneBefore}{PruneBefore}
\SetKwFunction{UpdateSubtree}{UpdateSubtree}

\Input{$\mathbf{x}_{start}$: New starting pose \\
       $B$: Budget}

$n_{match} \leftarrow$ \FindMatchingNode{$\mathbf{x}_{start}$}\;

\If{$n_{match}$ exists}
{
    \PruneBefore{$n_{match}$}\;
    \UpdateSubtree{$n_{match}$, $B$}\;
    $n_{start} \leftarrow n_{match}$
}
\Else
{
    $I_s \leftarrow I(\mathbf{x}_{start})$; $C \leftarrow 0$\;
    $n_{start} \leftarrow \{\mathbf{x}_{start}, I_s, C\}$\;
    $V \leftarrow {n_{start}}$; $E \leftarrow \emptyset$; $V_{closed} \leftarrow \emptyset$; $G \leftarrow (V,E)$\;
}
\caption{\textsc{UpdateGraph}}
\label{alg:updategraph}
\end{algorithm}

The Algorithm \ref{alg:ours1} described above is a planning procedure that generates a path from the starting configuration. To replan adaptively during flight, a starting configuration is chosen that is time $T$ in the future along the previously-planned path. Similarly, the budget $B$ is updated based on remaining flight time based on the current battery levels or mission constraints. The new path is merged with the previous path, keeping any portion of the previous path before the starting pose of the new path. This process is visualized in Fig.~\ref{fig:planner_replan}.

\subsection{Incremental Planning}\label{sec:incr-adap-plan}

Rebuilding the planning graph from scratch for every replan discards valuable information. In contrast, an incremental approach allows the planner to reuse previous planning efforts. Many scenarios, such as large budgets or small areas relative to the budget, can benefit from incremental refinement. The reuse of previous planning efforts not only improves the planner efficiency but also allows each new global plan to build on and refine earlier ones. In challenging planning situations, this approach allows the planner to iteratively expand the search tree, continuing to explore and evaluate potential paths over time. Even when the budget and belief map remain unchanged between replans, the previous graph remains valuable as it has been validated for collisions and adheres to the vehicle's motion model, allowing the planner to build upon prior work and save computational resources while maintaining accuracy.

\begin{figure*}[t]
\centering
\includegraphics[trim={0cm 5cm 0cm 0cm},clip,width=.99\textwidth]{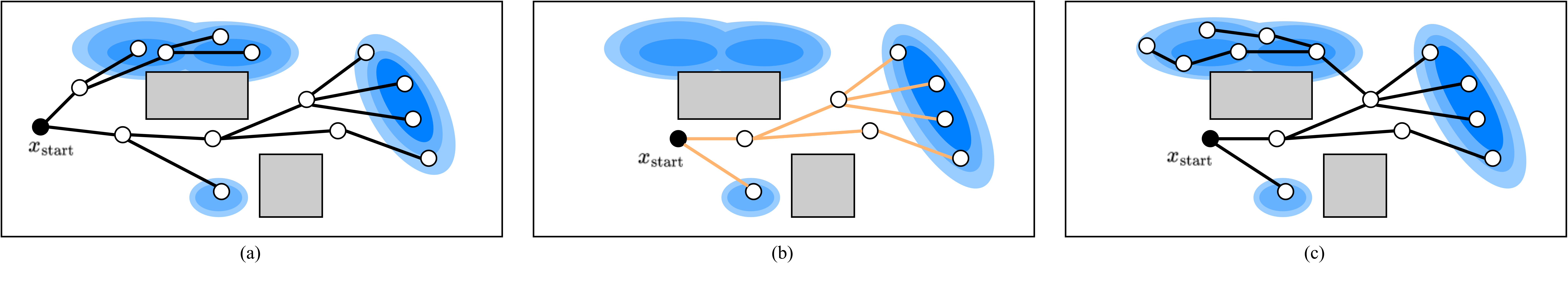}
\caption{An illustration of the incremental planning and refinement process in \PlannerName. (a) The global planner first generates long-horizon paths to maximize predicted information gain. The blue ellipses correspond to areas of information where areas with a larger alpha correspond to higher information. (b) At the start of each new planning cycle, infeasible portions of the tree, based on the executed trajectory and the current plan, are pruned. The remaining tree is then updated by recomputing information gain and cost based on the latest belief state and available budget, which is significantly more efficient than rebuilding the tree from scratch. (c) The tree is further expanded and refined for the duration of the planning cycle, after which the process repeats.
}
\label{fig:algorithm-breakdown}
\end{figure*}

To reuse the previous tree for incremental planning, \mbox{Algorithm~\ref{alg:ours1}} first checks whether the vertex set $V$ is empty. If it is not empty, rather than initializing the graph, \mbox{Algorithm~\ref{alg:ours1}} calls the function \textsc{UpdateGraph}, outlined in Algorithm~\ref{alg:updategraph}. This function first finds where the new starting point for the next plan matches a node in the existing previous trajectory (Line 1). When a matching node is found, all nodes and edges before that node are pruned away (Lines 2--3). The remaining portions of the graph are updated by checking the budget constraint and recomputing the information gain up to each node based on the current belief map. This is done efficiently by recursively traversing the tree using the helper function \textsc{UpdateSubtree} (Line 4), starting from the matching node and working down to all leaf nodes. The planner then uses the updated graph to continue planning as normal. If no matching node is found, the planner simply initializes a new graph from the new start pose (Lines 7--10). A high-level overview of this process is shown in Fig.~\ref{fig:algorithm-breakdown}.

\subsection{Belief Map Representation and Node Embedding}\label{sec:beliefspace}

In IPP, the information within a space can be represented using various models, such as Gaussian mixture models, Gaussian process regressions, grids, or voxels. This belief map can encode different types of information. For example, it might represent the probability of an object of interest being located in specific regions of the space. The choice of belief map formulation is important, as it directly influences the planning process by guiding search efforts and behaviors. An effective representation for a specific system, application, and planner balances computational complexity, memory requirements, and ability to perform operations on the belief map.

In this work, we adopt a grid-based representation of the belief map. Specifically, the search space is discretized into a grid of cells, where each cell $i$ is associated with a binary random variable $X_i$ indicating the presence ($X_i=1$) or absence ($X_i=0$) of an object of interest. The probability $P(X_i)$ represents our belief about an object being located in cell $i$. This grid-based representation is computationally straightforward to update with new or predicted observations using a Bayesian framework and an appropriate sensor model.

While we adopt a binary occupancy grid here to represent the presence of discrete objects, other belief models can capture spatial correlations and are better suited for continuous-valued fields like temperature or gas concentration. These richer models, however, typically increase computational cost. In our target applications, where objects of interest are assumed to be independent and onboard real-time planning is critical, the grid-based model is a practical choice. Importantly, our planner itself is agnostic to the belief representation and could be adapted to different models as needed.

\subsubsection{Belief Map Node Embedding}\label{sec:embedding}

The nodes in the planning tree maintained by the IA-TIGRIS algorithm store the necessary information to represent the evolving belief map. 
For our implementation, the information gain up to a given node is computed based on the change in entropy within the grid cells observed by the agent’s sensor footprint during traversal from the root node to the current node.

A straightforward approach to calculating the information gain for a given node consists of traversing the tree from the root all the way to the given node. It is necessary to evaluate the entire trajectory because the information gain can be not modular and previous observations affect the information gain of future observations. 
Calculating the additional information gain contributed by a new node and edge requires knowledge of the state of the belief map at the previous node. However, this approach of calculating information gain starting at the root node is computationally inefficient and slow.

Let $n$ denote the number of grid cells in the belief map, $m$ the total number of nodes in the planning tree, $k$ the number of nodes in the trajectory from the root to the current node, and $j$ an upper bound on the number of cells affected by a single sensor footprint. Computing the information gain for the current node entails stepping through each of the $k$ nodes along the trajectory and, at every step, updating at most $j$ cells, giving a running time of $\mathcal{O}(kj)$. Because the belief map is maintained once and updated in place during this traversal, the memory footprint remains $\mathcal{O}(n)$.

One approach to speeding up computation is to store the entire belief map within each node. In this approach, the information gain for a new node is simply the sum of the parent node’s information gain and the contribution of the new node.  With this design, evaluating a newly added node involves touching only the $j$ cells observed at that node, so the per-update cost drops to $\mathcal{O}(j)$. The trade-off is a higher space requirement: each of the $m$ nodes now holds $n$ cell probabilities, for a total memory complexity of $\mathcal{O}(mn)$. While feasible for small maps, this method becomes impractical for large-scale maps relevant to the domains and applications considered in this work due to its high memory requirements.

To efficiently compute the information gain for a new node, we introduce a node embedding strategy that eliminates the need for a full traversal of the planning tree from the root to the current node, by embedding only the delta of the belief state at each node. Concretely, every node stores the subset of grid map cells---no more than $j$ of them---that its sensor footprint updates. This subset is stored as a hash map, keyed by cell index, maintaining the most recent probability of the visited cell; unvisited cells fall back to the baseline grid that is held once for the whole problem. When a new node is appended, the planner consults the hash map for at most $j$ entries, updates those values, and aggregates the resulting change in entropy into the node’s cumulative information-gain value, so the per-node running time is $\mathcal{O}(j)$.

Because each node retains at most $j$ key–value pairs, the additional storage it contributes is $\mathcal{O}(j)$, and the tree as a whole adds $\mathcal{O}(jm)$ to the single global grid of size $n$. Provided $j \ll n$, the combined memory footprint, $\mathcal{O}(jm + n)$, is therefore orders of magnitude smaller than the $\mathcal{O}(mn)$ required by the naive approach while achieving the same incremental information-gain calculation.

While implementing this embedded method, we observed that hash map key selection significantly impacts computation time. Using strings as keys resulted in substantial overhead due to frequent string-to-integer conversions. Even transitioning from a two-integer key to a single-integer key yielded a notable improvement in performance and is the approach we use in our implementation through a simple index-to-key conversion. Since cell value lookups occur frequently, optimizing this efficiency significantly accelerated search tree expansion.

By incorporating this belief map node embedding strategy, our planning framework efficiently maintains an accurate representation of the evolving belief map within the search tree. This enables the planner to greatly reduce computational overhead and create higher quality paths. 

\subsection{Sensor Modeling}\label{sec:sensor-model}

The updates to our belief map are dependent on the sensor model, which captures the performance characteristics of the perception system employed. There are many environmental factors that affect the performance of a sensor, such as visibility and range, which can be incorporated into this model \cite{thrun2000probabilistic, Carbone_2022, popovic2020informative}. 

For our approach, we formulate the sensor model as a piecewise function that is implemented as a lookup table. For a given range $r$ and binary measurement $Z$, this table maps the range of the object from the sensor to the corresponding true positive rate (TPR) $P\left(Z|X,r\right)$ and true negative rate (TNR) $P\left(\overline{Z}|\overline{X},r \right)$ of an object detection model. A typical range-based detection model will have better performance at close ranges and then taper off as the distance increases. Using the TPR and TNR of the sensor model, we apply Bayes' theorem to update the posterior belief of $X_i$ in the belief map given the measurement $Z$.

By maintaining a lookup table of these sensor models, we can easily incorporate different sensor modalities and their corresponding capabilities into the planning framework. The sensor model can be derived from actual test data and easily integrated into the framework rather than having to fit a function to the results.

The sensor model for belief map updates is a crucial component of the overall planning framework, as it directly affects the information gain computed for each node in the IA-TIGRIS tree. Accurately modeling the sensor’s performance characteristics enables the planner to make informed decisions, resulting in more effective exploration and search strategies that align with actual sensor capabilities. Conversely, inaccuracies in the sensor model can lead to overconfidence in observed areas or unnecessary revisits to regions that have already been sufficiently searched.

\subsection{Information Reward Models}\label{sec:reward-models}

The information reward function $I(\cdot)$ used in our framework is not constrained to being modular and could be time-varying or even submodular, depending on the specific requirements of the application. In this work, we have implemented an objective of reducing the overall entropy of the belief map. As defined in Section \ref{sec:beliefspace}, our belief map represents the probability of the presence of an object of interest in a cell. The reward for one cell $I(X_i)$ is calculated by finding the difference between the initial entropy and the final entropy after updating the belief with the measurements and sensor model defined in Section \ref{sec:sensor-model}.

The Shannon entropy of a cell is calculated by 
\begin{equation}
    H(X_i) = -P(X_i)\log P(X_i)-P(\overline{X}_i) \log P(\overline{X}_i)
    \label{shannon}
\end{equation}
where $X_i$ represents the event that the cell is occupied and $\overline{X}_i$ denotes its complement, the event that the cell is free.

The entropy reduction due to a single positive measurement would be calculated as
\begin{equation*} %
\Delta H(X_i|Z) = H(X_i) - H(X_i|Z)
\end{equation*}
where $H(X_i)$ is given by (\ref{shannon}), and $H(X_i|Z)$ is the entropy conditioned on the sensor measurement $Z$. 

The final information gain for a trajectory $I(\mathcal{T})$ would be
\begin{equation*}
    I(\mathcal{T}) = \sum_i^I H(X_{i,init}) - H(X_{i,final})
\end{equation*}
where $X_{i,init}$ is the initial belief state of cell $i$, and $X_{i,final}$ is the final belief state.

\subsubsection{Information Reward Estimation}\label{info-reward}

\begin{figure}[t]
\centering
\includegraphics[trim={5.5cm 0cm 0cm 3.5cm},clip,width=0.99\columnwidth]{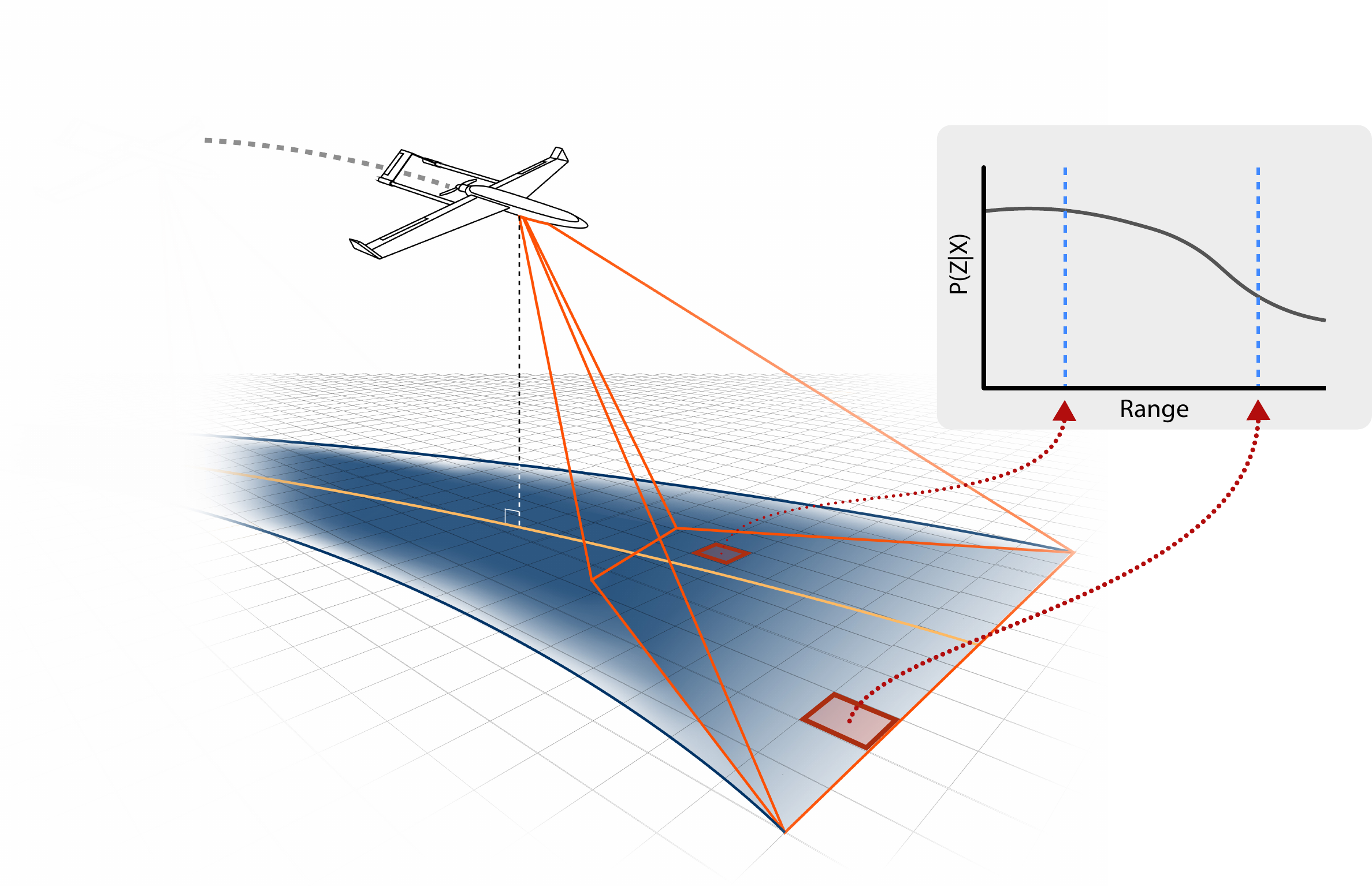}
\caption{An example of how the grid cells rewards are estimated given a planned trajectory. The cells that are closer to the center of the trajectory have a smaller minimum distance than cells farther from the center of the trajectory, leading to a larger change in entropy for the closer cells. 
}
\label{fig:sensor-model}
\end{figure}
When a new node is added in the planning tree, the approach for estimating the information gain may vary depending on the modularity and temporal dependence of the information function. In the case of a modular and temporally independent function, the reward for a new node can be simply appended to the cumulative reward along the path. Alternatively, if the information function exhibits non-modularity or temporal dependence, the entire trajectory up to the current node may need to be reevaluated to determine the appropriate reward. 
Because we do not yet know the future measurements for each cell, the information gain for the planned path must be estimated. 

One option would be to compute the expectation of the information gain given all possible future measurements. However, this approach can be computational expensive and infeasible. Rather than compute the expectation, we follow the approach of \cite{hollinger_long-horizon_2015,moon2022tigris} and use an optimistic approximation of the expected reward. Specifically, we assume a positive measurement $Z$ if $P(X) \geq 0.5$, and a negative measurement otherwise. This assumption leads to the following cell reward function for a single measurement:
\begin{equation}
    I(X_i) = \begin{cases} \Delta H\left(X_i|Z\right) & P\left(X_i\right) \geq 0.5 \\ \Delta H\left(X_i|\overline{Z}\right) & P\left(X_i\right) < 0.5 \end{cases}
    \label{equ:reward}
\end{equation}

To compute the information reward along a path, we project the sensor footprint onto the surface plane to determine all the cells it contains. The reward for each cell is calculated using \eqref{equ:reward} and then aggregated. This process becomes more intricate when incorporating graph edges into the reward computation. Discretizing the trajectory between nodes into individual views creates imperfect overlaps due to the non-square sensor footprint, potentially omitting some grid cells from the edge reward calculation. Mitigating this issue requires significant overlap between sensor footprints, which, in turn, greatly increases the computation time for updating grid cell beliefs.
Rather than selecting a discretization resolution for the trajectory, we compute a conservative approximation of the edge reward by updating all cells within the view of edge trajectory, using the closest viewing distance along the path for each cell.
This approach ensures that all cells along the edge are accounted for, with each belief value updated using a measurement taken when the sensor is closest to that cell. A visualization of this reward estimation is shown in Fig. \ref{fig:sensor-model}.

\subsubsection{Priority-Dependent Rewards}\label{sec:priority}
In the case where certain areas of an environment or spatial information are more important, such as areas around populated regions or critical infrastructure, we introduce the ability to weight the information reward by a priority scalar, $p_i$, representing the priority of cell $i$. This approach provides finer control over the UAV behavior by allowing the entropy of different cells to be weighted unequally. Assigning a higher priority to some cells over others results in paths that favor observations of the higher-weighted cells. This capability is also useful when specific instances of an object are more important than others, and the spatial distribution of those instances differs from the overall class distribution in the search space (as demonstrated in Section \ref{sec:priority-time-eval}).

\subsubsection{Time-Dependent Rewards}
The IPP formulation maximizes total information gain over the entire mission without explicitly relating the reward to the timing of observations---rewards are time-independent. However, in some applications, operators may prefer prioritizing areas of high information gain earlier in the trajectory rather than later.

\begin{figure}[t]
\centering
\includegraphics[width=\columnwidth]{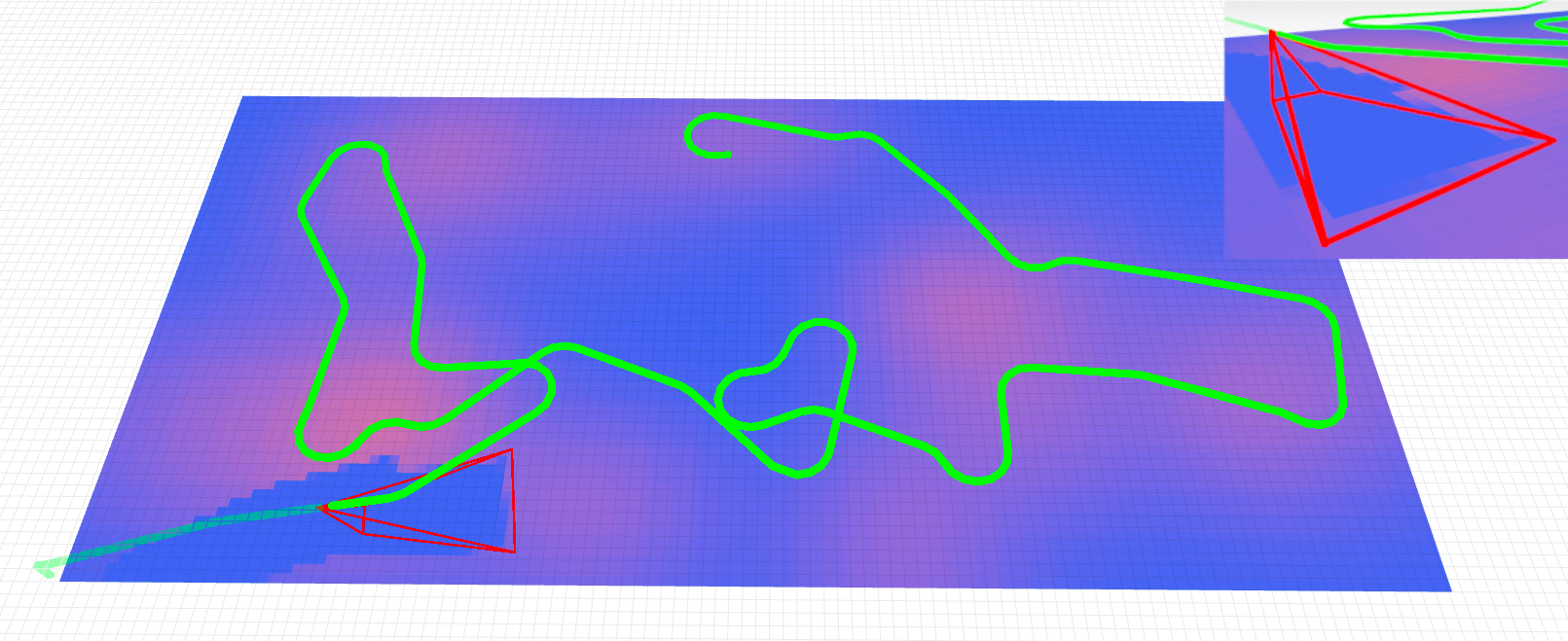}
\caption{Example results for IA-TIGRIS after a replanning step in our simulation environment. Pink regions represent regions of high entropy while blue regions are areas of low entropy. The red projection shows the camera frustum while the green line represents the planned path.}
\label{fig:simple_sim}
\end{figure}

We introduce the ability to create time-dependent reward that decreases the reward value for all cells over the mission duration, resulting in plans that prioritize visits to areas of high information gain early in the plan. This capability works especially well when paired with priority-dependent rewards, as introduced in Section \ref{sec:priority}. The UAV will be biased toward visiting the higher-priority priors early in the plan, and the operator can even roughly dictate the order to visit areas through differing priority weights and using the time-dependent reward decay function.

In our use cases, we implement a simple decay function
\begin{equation*}
    \Gamma(t) = \begin{cases}  \gamma & t \geq \gamma_t \\ 
                               \beta t + 1 & t < \gamma_t 
                               \end{cases}
\end{equation*}
where $\Gamma(t) \in [\gamma, 1]$, and $\gamma \in [0, 1]$ is the floor value of the decay function. $\beta < 0$ is the decay rate and $\gamma_t = (\gamma-1)/\beta$ is the time where the function output becomes the floor value $\gamma$. This function results in a linear decay until the floor value. 

Combining our decay function $\Gamma(t)$ and cell priority values $p_i$ creates a priority- and time-dependent reward function 
\begin{equation*}
    I(X_i,t) = \begin{cases}  p_i\Gamma(t)\Delta H\left(X_i|Z_i\right) & P(X_i) \geq 0.5 \\ p_i\Gamma(t)\Delta H\left(X_i|\overline{Z}_i\right) & P(X_i) < 0.5 \end{cases}
\end{equation*}

%% file: tex/4_evaluation.tex
\section{Simulation Evaluation \& Results}\label{sec:results}

To rigorously evaluate the performance and benefits of \PlannerName, we first conduct tests in a simulation environment. These experiments include ablations to assess the impact of key design decisions, as well as Monte Carlo evaluations comparing \PlannerNameSpaced against baseline methods. 
The codebase for our planner, simulation environment, and testing package can be found at 
\href{https://ia-tigris.github.io}{https://ia-tigris.github.io}.

\begin{figure*}[t]
\centering
\begin{tikzpicture}
    \node[anchor=south west, inner sep=0] (image) at (0,0) {
        \includegraphics[trim={0.0cm 2.5cm 0.0cm 3.5cm},clip,width=.99\textwidth]{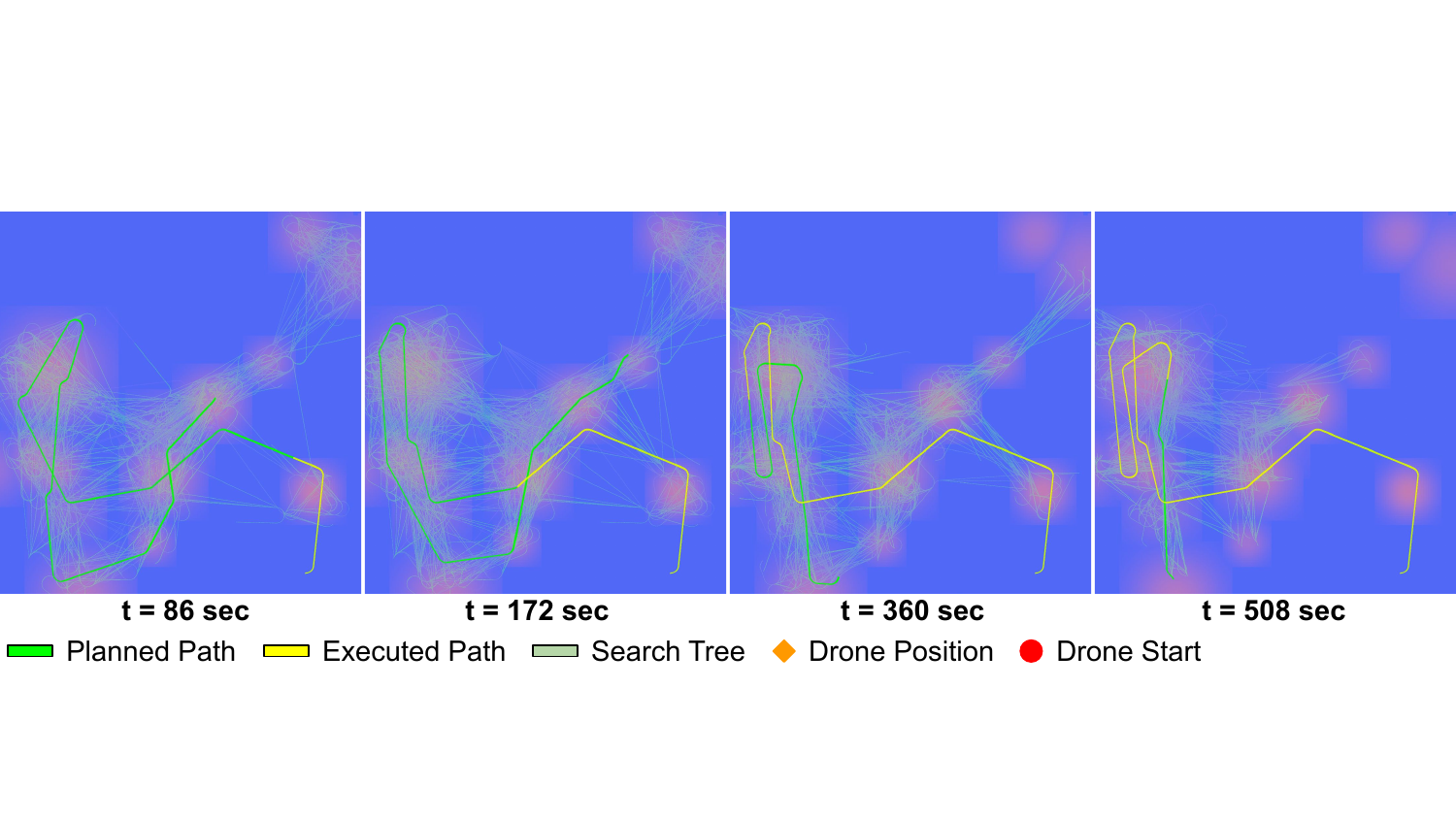}
    };
    \begin{scope}[x={(image.south east)},y={(image.north west)}]
        \fill[OrangeRed] (0.21, 0.218) circle (1.5pt); 
        \fill[OrangeRed] (0.46, 0.218) circle (1.5pt); 
        \fill[OrangeRed] (0.71, 0.218) circle (1.5pt); 
        \fill[OrangeRed] (0.96, 0.218) circle (1.5pt);

        \node[diamond, fill=orange, scale=0.30] at (0.202, 0.458) {};
        \node[diamond, fill=orange, scale=0.30] at (0.355, 0.398) {};
        \node[diamond, fill=orange, scale=0.30] at (0.5145, 0.581) {};
        \node[diamond, fill=orange, scale=0.30] at (0.801, 0.62) {};
    \end{scope}
\end{tikzpicture}
\caption{IA-TIGRIS continuously replans and refines the global path over the simulation. These snapshots show how the path adapts online to maximize information gain, with the search tree visualized to illustrate focus on high-information areas while respecting the remaining budget. For clarity, the visualized entropy map is static in the figure to highlight the evolving planned path, but in reality the belief map updates continuously as measurements are taken.
}
\label{fig:example_run}
\end{figure*}

\subsection{Simulation Framework}\label{sec:sim-framework}

We use a simplified simulator, shown in Fig.~\ref{fig:simple_sim}, to enable rapid development and validation of our approach. This simulator models scenarios where a UAV is tasked with mapping the location of objects of interest in the belief map by planning paths that maximize the reduction of the entropy. The prior belief map could be informed by previous flight data, user input, or data from other sensor modalities. The visualized color for each grid cell represents the entropy of the belief that an object lies within that space, with pink regions corresponding to areas of higher entropy and blue regions indicating low entropy. The simulation employs a constant-velocity motion model for the fixed-wing UAV. To account for the effects of banking on observations, we compute the orientation change between consecutive waypoints. If this change exceeds a threshold, the UAV is considered to be banking, and the belief map is not updated during this period. This consideration reflects the fact that the camera is primarily pointed toward the sky during banking and is also affected by increased motion blur.

When the UAV reaches a commanded waypoint within a specified Euclidean distance threshold, the next waypoint in the list becomes the new commanded waypoint. During replanning, a newly generated path is merged with the existing path, incorporating the updated trajectory starting from the beginning of the new path. A simple proportional (P) controller is used to regulate heading and altitude, guiding the UAV between waypoints.

The UAV commanded speed is $25$ m/s, with a planning time of $10$ seconds and a flight and sampling altitude of $50$ m. The turning radius is $100$ m, and the onboard camera is pitched $30$ degrees from horizontal with a $36.9$-degree field of view. The TPR and TNR values are $0.9$ for ranges of $0$--$200$ m and then decrease linearly to $0.5$ at $600$ m. The belief map grid cells are $30$ m, and the total budget for flight distance is $15,000$ m. The belief map is updated using Bayes' theorem to incorporate the projected camera field of view, simulated observations, and the sensor model. To accelerate testing, the simulation runs at twice the real-time speed, with results subsequently rescaled to match real-time conditions.

Fig.~\ref{fig:example_run} presents four snapshots from a simulation. The planned path is shown in green, the executed path in yellow, the robot's current position as an orange diamond, and the starting position as a red circle. Early in the simulation, the planned trajectory makes two passes over the larger clusters of high-entropy regions, which represent areas of high uncertainty. The search tree explores a wide range of paths across these regions to maximize information gain, refining its evaluation over time. As the planning cycle progresses, the search tree is iteratively updated through tree recycling, which improves computational efficiency by retaining and modifying previous solutions rather than replanning from scratch. At $t=180$, the plan adapts by doubling back to the middle-left region, due to finding a more rewarding path. Toward the end of the simulation, the effect of the budget constraint becomes apparent, as the search tree no longer extends into the upper-right corner of the search area.

\subsection{Metrics}

The primary metric for evaluating the success of our plan is the information gain along the executed path, quantified as the change in total entropy of the belief map from the initial to the final state. To facilitate interpretation and comparison across tests, we report this change as the percent reduction in entropy. Initially, the entropy reduction is zero, as no observations have been made. As the robot gathers information, the entropy decreases, reflecting an improved understanding of the environment. This reduction accumulates during the simulation as new observations refine the belief map.

\subsection{IA-TIGRIS Ablations}

We conduct ablation experiments to systematically evaluate the contributions of incremental and adaptive planning, belief map node embeddings, priority- and time-dependent rewards, and planner parameters.

\subsubsection{Incremental and Adaptive Planning Evaluation}

We first evaluate the efficiency of recycling a search tree compared to building a new one for each planning cycle. For \mbox{IA-TIGRIS} to be both adaptive and incremental, reusing and recycling previous planning efforts must be computationally efficient. As explained in Section \ref{sec:incr-adap-plan}, each planning cycle begins by pruning irrelevant trajectories, checking budget constraints for remaining nodes, and updating information gain based on the current belief map. If these updates consume too much planning time, there will be insufficient time left to refine and extend the trajectory.

Our implementation of IA-TIGRIS demonstrates that recycling a previously built tree is significantly more efficient than constructing a new tree from scratch. For example, a tree that takes $10$ seconds to build requires less than $0.2$ seconds to update under our planner settings and test environment. This efficiency gain allows the entire tree to be recycled and extended with the remaining planning time, even if none of the previous plan was executed. By avoiding the computational overhead of trajectory generation, neighborhood search, and collision checking, tree recycling results in faster updates and enables the planner to generate improved paths within the same time constraints.

To quantify the impact of tree recycling and replanning, we conducted 50 tests in the simulation framework described in Section \ref{sec:sim-framework}. Without replanning, the average reduction in entropy was $20.4\!\pm\!2.8$\% (mean $\pm$ 95\% CI), which would be close to results expected from \cite{moon2022tigris}. Introducing replanning without tree recycling improved this reduction to $23.1\!\pm\!3.3$\%, a $13.2$\% relative increase in performance. Even in the absence of environmental disturbances, replanning improves path quality by addressing mismatches between estimated and actual information gain, particularly as the planning horizon shortens due to decreasing budget. Incorporating incremental planning with tree recycling further increased the reduction in entropy to $25.4\!\pm\!3.8$\%, a 24.5\% relative improvement compared to replanning without tree recycling and a statistically significant difference ($p\!<\!0.05$). This result highlights \PlannerName's ability to refine and improve the previous global plan.

\begin{figure}[t]
    \centering
    \includegraphics[trim={0.4cm 0.4cm 0.4cm 0.3cm},clip,width=0.95\columnwidth]{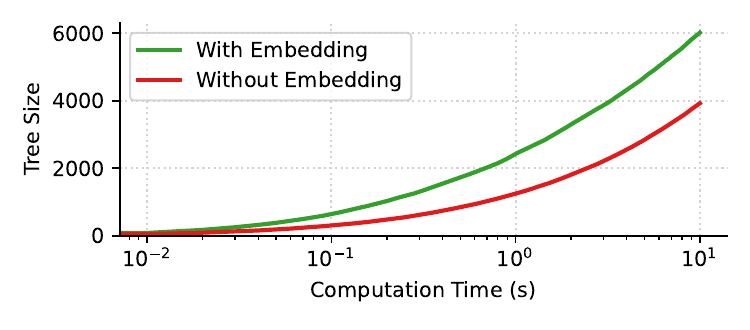}
    \caption{Tree size vs. computation time for 100 trials.}
    \label{fig:node-embedding}
\end{figure}

As expected, the benefits of replanning and refinement become more pronounced when planning time is reduced. When the allowed planning time is reduced to 3 seconds, the relative performance difference between no replanning and replanning with tree recycling increases to 25.8\%. This gap would widen further for even shorter planning times, where initial plans tend to be of lower quality. 
However, as planning time decreases, a larger portion of the time is allocated to recycling previous plans rather than expanding the tree, leaving less time for refinement. The planning time inherently introduces a tradeoff between path quality and responsiveness; more frequent replanning improves adaptability to new information, while longer planning cycles allow for more extensive trajectory refinement. In the majority of our experiments and applications, we set a planning time of 10 seconds, but the exact planning frequency depends on scenario complexity and computational resources.

\subsubsection{Belief Map Node Embedding Analysis}

We evaluated the impact of belief map embedding on planning efficiency. Calculating the information gain for a trajectory can take up a disproportionate amount of the planning time if the belief map is not implemented well. We found that in our typical testing environments, using our belief map embedding in the tree nodes reduced the time to compute the information gain for each node by around $80$--$95\%$. This increase in speed only resulted in an additional ${\sim}20$ MB of memory usage for the search tree.

The cumulative effect of this speedup is demonstrated in Fig.~\ref{fig:node-embedding}, which compares tree size and computation time for the first planning iteration, with and without embeddings, across 100 trials on random simulation environments. Due to the reduction in time to add each new node to the planning tree, the tree size is much larger at any point during planning. 
On the initial planning cycle, approximately $84\%$ of the computation in each planning loop is spent on information gain evaluation, and most of the remaining $16\%$ is devoted to the steering function (of which roughly $3\%$ involves collision checking). This distribution highlights why accelerating information gain evaluation has such a large effect on increasing the rate at which new nodes can be added to the trajectory tree.

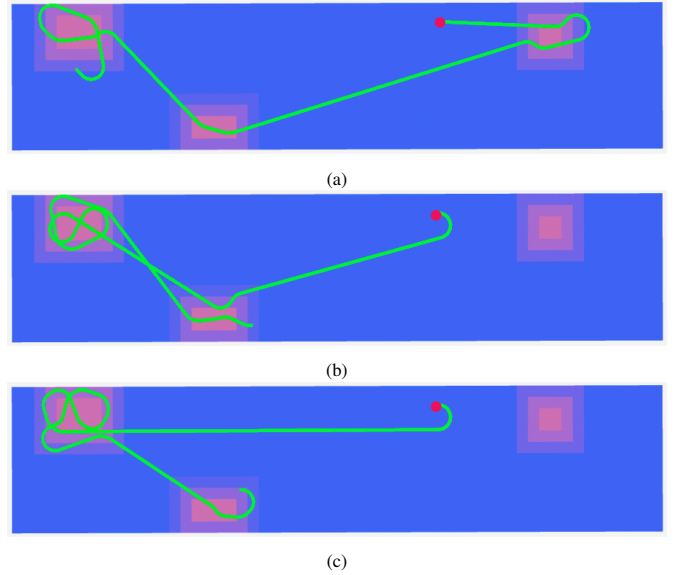
\begin{figure}[t]
    \centering
    \begin{subfigure}[b]{0.99\columnwidth}
        \centering
        \begin{tikzpicture}
            \node[anchor=south west, inner sep=0] (image) at (0,0) {
                \includegraphics[trim={0cm 0.0cm 0cm 0cm},clip,width=\columnwidth]{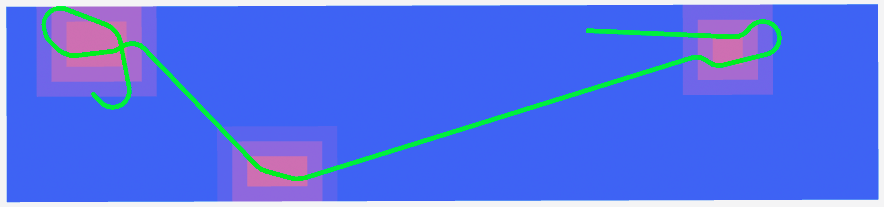}
            };
            \begin{scope}[x={(image.south east)},y={(image.north west)}]
                \fill[OrangeRed] (0.656, 0.85) circle (2pt); %
            \end{scope}
        \end{tikzpicture}
        \caption{}
        \label{fig:priority_time_impacta}
    \end{subfigure}
    
    \begin{subfigure}[b]{0.99\columnwidth}
        \centering
        \begin{tikzpicture}
            \node[anchor=south west, inner sep=0] (image) at (0,0) {
                \includegraphics[trim={0cm 0cm 0cm 0cm},clip,width=\columnwidth]{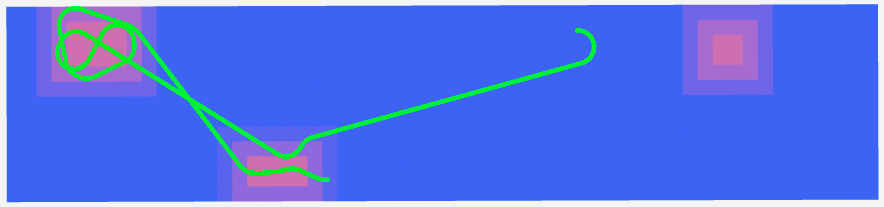}
            };
            \begin{scope}[x={(image.south east)},y={(image.north west)}]
                \fill[OrangeRed] (0.65, 0.84) circle (2pt); %
            \end{scope}
        \end{tikzpicture}
        \caption{}
        \label{fig:priority_time_impactb}
    \end{subfigure}

    \begin{subfigure}[b]{0.99\columnwidth}
        \centering
        \begin{tikzpicture}
            \node[anchor=south west, inner sep=0] (image) at (0,0) {
                \includegraphics[trim={0cm 0cm 0cm 0cm},clip,width=\columnwidth]{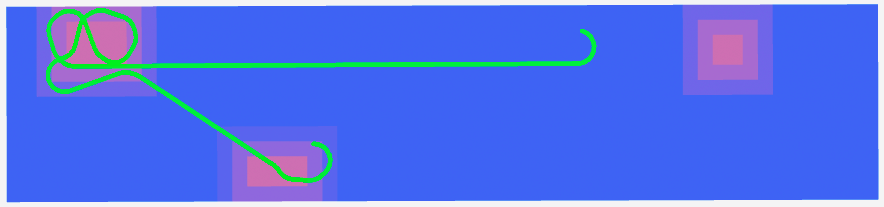}
            };
            \begin{scope}[x={(image.south east)},y={(image.north west)}]
                \fill[OrangeRed] (0.65, 0.84) circle (2pt); %
            \end{scope}
        \end{tikzpicture}
        \caption{}
        \label{fig:priority_time_impactc}
    \end{subfigure}

    \caption{Examples showing how modifying the priority- and time-based rewards affects behavior. The starting point is shown in red and the planned path in green. (a) The base case with equal priorities. (b) A higher priority on the far-left prior causes the drone to spend more of its budget observing that space. (c) With time-based rewards, the drone visits the high-priority prior as quickly as possible to maximize reward.}
    \label{fig:priority_time_impact}
\end{figure}

Our embedding also has a profound impact on how quickly the search tree is recycled between plans. 
Previously the information gain at each node would be calculated individually when the tree is recycled, with the nodes toward the base of the tree getting evaluated many times. 
Now, the tree can be updated recursively from the root to the leaves without the information at any node being evaluated more than once. 
In our typical environments, this optimization reduced the per-node tree rebuild time from 
$180.93~\upmu\text{s}~(\sigma = 14.7~\upmu\text{s})$
to $15.48~\upmu\text{s}~(\sigma = 1.9~\upmu\text{s})$.

We also evaluate the impact of belief map embedding by using Google Benchmark \cite{gtest} on an extremely large environment. Google Benchmark runs each test repeatedly until the results are statistically stable, automatically accounting for noise and warm-up effects. For these benchmark tests, the number of nodes in the trajectory was $k = 7$, the number of grid cells in the search space was $n=39,841,344$, and the number of grid cells updated by the new node was $j=11,984$. The time to evaluate the information gain for a new node decreased from $43$ ms to $1.9$ ms, a $96\%$ improvement. This decrease led to the size of the search trees to on average be more than double in size.

\begin{figure}[t]
    \centering

    \begin{tikzpicture}
        \node[anchor=south west, inner sep=0] (image) at (0,0) {\includegraphics[trim={1cm 0.7cm 1.4cm 1cm},clip,width=\columnwidth]{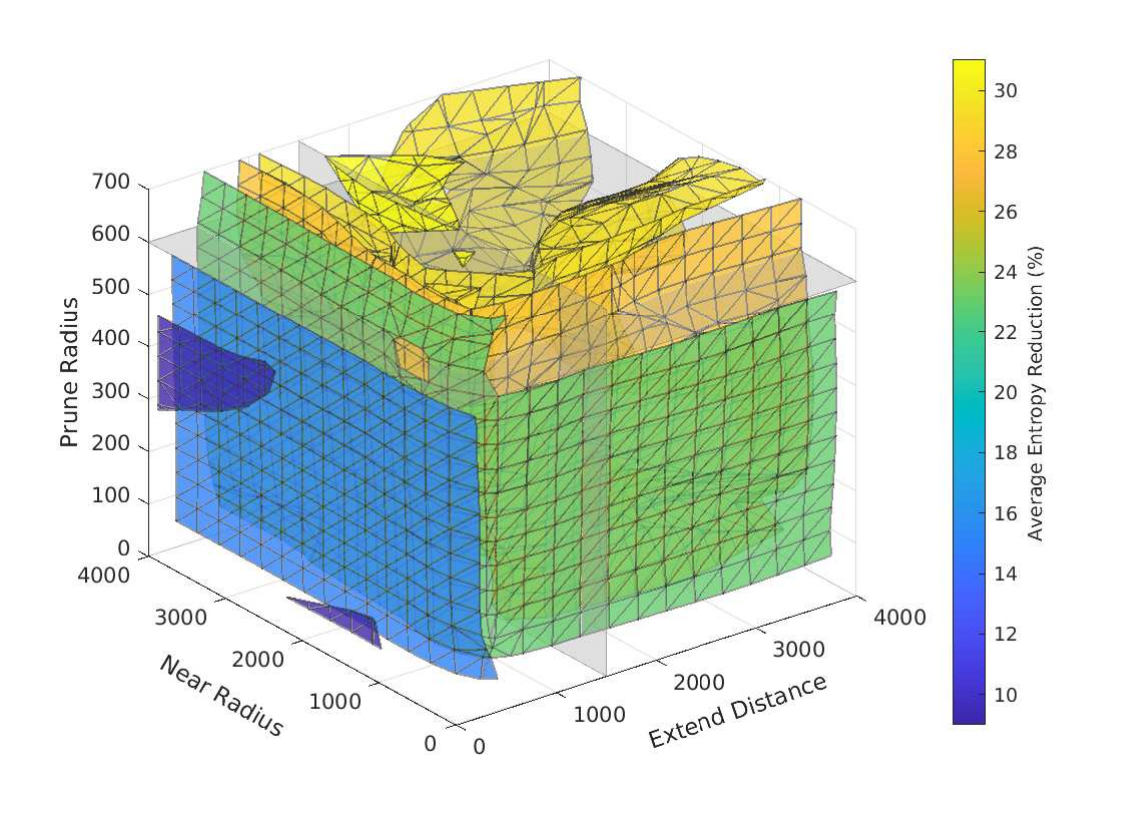}};
        
        \node[anchor=north west] at (4.75,1.25) {\scriptsize{(a)}};
        \node[anchor=north west] at (7,4.6) {{\scriptsize(b)}};
    \end{tikzpicture}

    \begin{subfigure}{0.48\columnwidth}
        \centering
        \includegraphics[trim={0.2cm 0.2cm 0.7cm 0.3cm},clip,width=\linewidth]{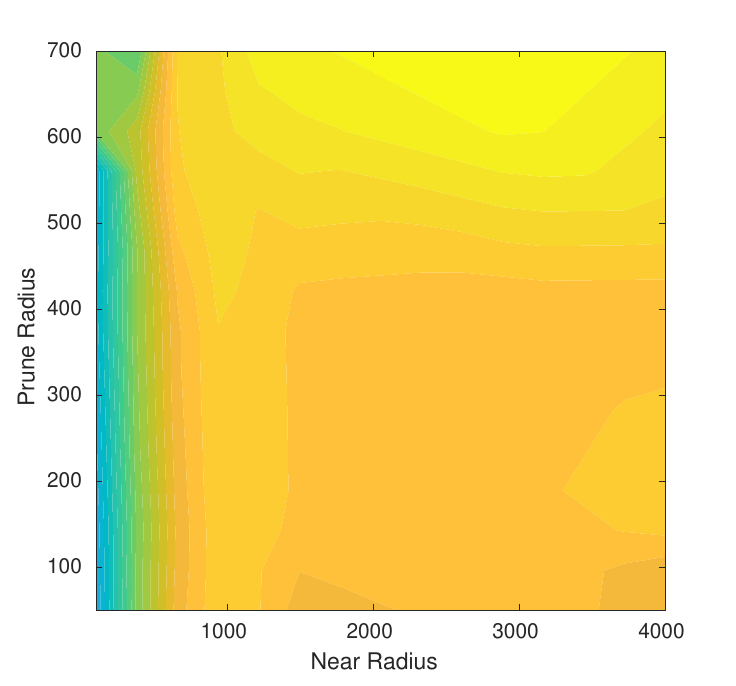}
        \caption{}
        \label{fig:slice-prune}
    \end{subfigure}
    \hfill
    \begin{subfigure}{0.48\columnwidth}
        \centering
        \includegraphics[trim={1.6cm 0.2cm 2.2cm 0.3cm},clip,width=\linewidth]{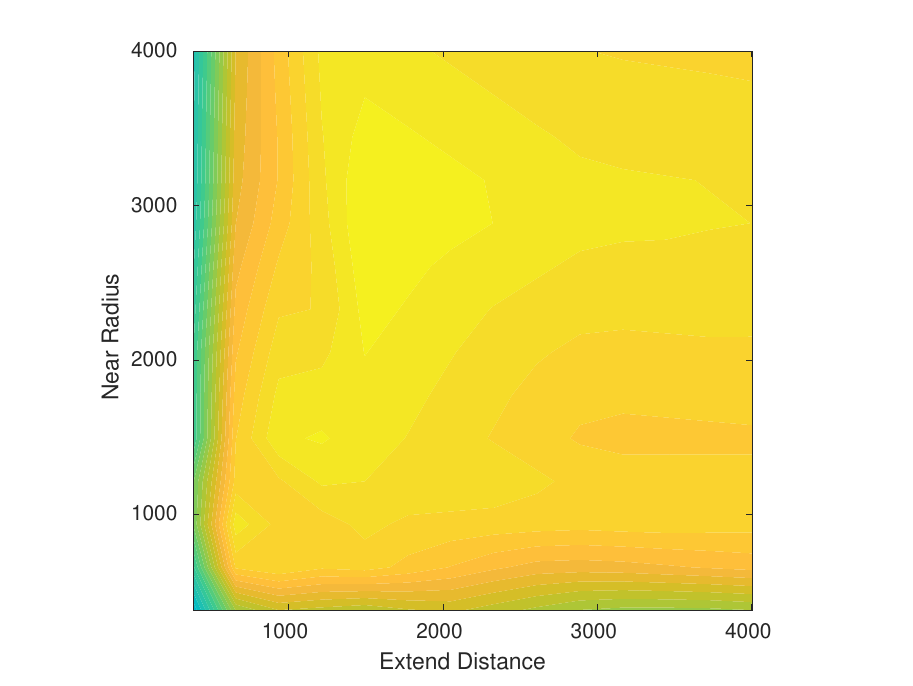}
        \caption{}
        \label{fig:slice-extend}
    \end{subfigure}

    \caption{Top: Isosurface plot showing global parameter sweep and resulting entropy reduction. Each surface shows given parameters that result in similar average percent of entropy reduction. Bottom: Slices of the data at fixed extend distance (a) and pruning radius (b).}
    \label{fig:param-sweep}
\end{figure}

\subsubsection{Priority- and Time-Dependent Rewards Evaluation}\label{sec:priority-time-eval}

To illustrate the impact of priority-dependent and time-dependent rewards, we conduct a simulation with the results shown in Fig.~\ref{fig:priority_time_impact}. To make the effect on flight paths clearer, we pitch the camera down so that the flight path coincides more closely with what is observed by the sensor. The example scenario has three clusters of uncertainty in the search space.

Fig.~\ref{fig:priority_time_impacta} shows the base case where three areas of uncertainty have equal priority. The planned path visits the closest cluster first and then travels to the left side, observing all three clusters with the sensor. Fig.~\ref{fig:priority_time_impactb} demonstrates when the cluster on the far left is adjusted to have a higher priority. The path no longer visits the cluster on the right in favor of saving the budget to expend more effort in searching the higher-priority region. Fig.~\ref{fig:priority_time_impactc} shows how the path first visits the high-priority prior before moving on to the other prior on the left when a time-based reward is added along with priority-based rewards. The time-based reward causes the order of observations to affect the total reward.

\subsubsection{Planner Global Parameter Search}
We investigate the effects of the extend distance $\Delta$, the near radius $R$ used in \textsc{Near}$(\cdot)$, and the pruning radius used in \textsc{Prune}$(\cdot)$ within \PlannerName, as outlined in Section \ref{sec:alg_overview}. To investigate these effects, we perform a global parameter search across 25 randomly generated environments. Each environment was 5 km $\times$ 5 km, with a prior belief map consisting of a random number of Gaussian distributions with varying locations, means, and standard deviations. The planning loop was given a $10$-second planning time, and the agent continued to observe the space until budget was expended.

Using the average percent reduction in entropy as the performance metric, we visualize the results in an isosurface plot, shown in Fig.~\ref{fig:param-sweep}, where higher percentages indicate greater information gain. To provide further insight, we include two 2D slices of the data at the bottom of the figure: one at an extend distance of 1500 m and the other at a pruning radius of 600 m. These slices reveal the structure of the solution space along the individual parameter dimensions.
The results suggest that, for the sampled environments, planner performance improves as the extend distance and near radius are larger. If one is much smaller than the other, performance is greatly reduced. However, the relationship between pruning radius and performance is less clear, though a slight performance increase is observed as the pruning radius increases.

The pruning radius primarily affects the planned path length at the start of a test. A very small pruning radius results in paths shorter than the $15$ km budget, likely because a denser tree will reduce tree depth. Similarly, reducing the extend distance below $500$ m leads to shorter initial plans, suggesting that both parameters play a role in determining tree expansion.

We find that as the extend distance increases, maintaining a sufficiently large near radius is crucial for sustaining performance. If the near radius drops below the extend distance, performance declines. Meanwhile, the pruning radius can be set relatively high without significant performance loss, and doing so allows for longer planned paths. Additionally, the CPU performance of a given system may influence the optimal parameter settings, suggesting the need for adjusting parameters based on computational resources.

The parameter with the highest performance in our testing had the combination of an extend distance and near radius that balanced expansion while still creating a dense tree, and a large pruning radius to act as a stronger heuristic for high-quality paths. For our subsequent simulation testing, we selected an extend distance of $1500$ m, near radius of $1500$ m, and a pruning radius of $600$ m.

\subsubsection{Long-Horizon Planning}

\begin{figure}[t]
\centering
\begin{tikzpicture}
    \node[anchor=south west, inner sep=0] (image) at (0,0) {
        \includegraphics[trim={0.25cm 0.8cm 0.25cm 1cm},clip,width=\columnwidth]{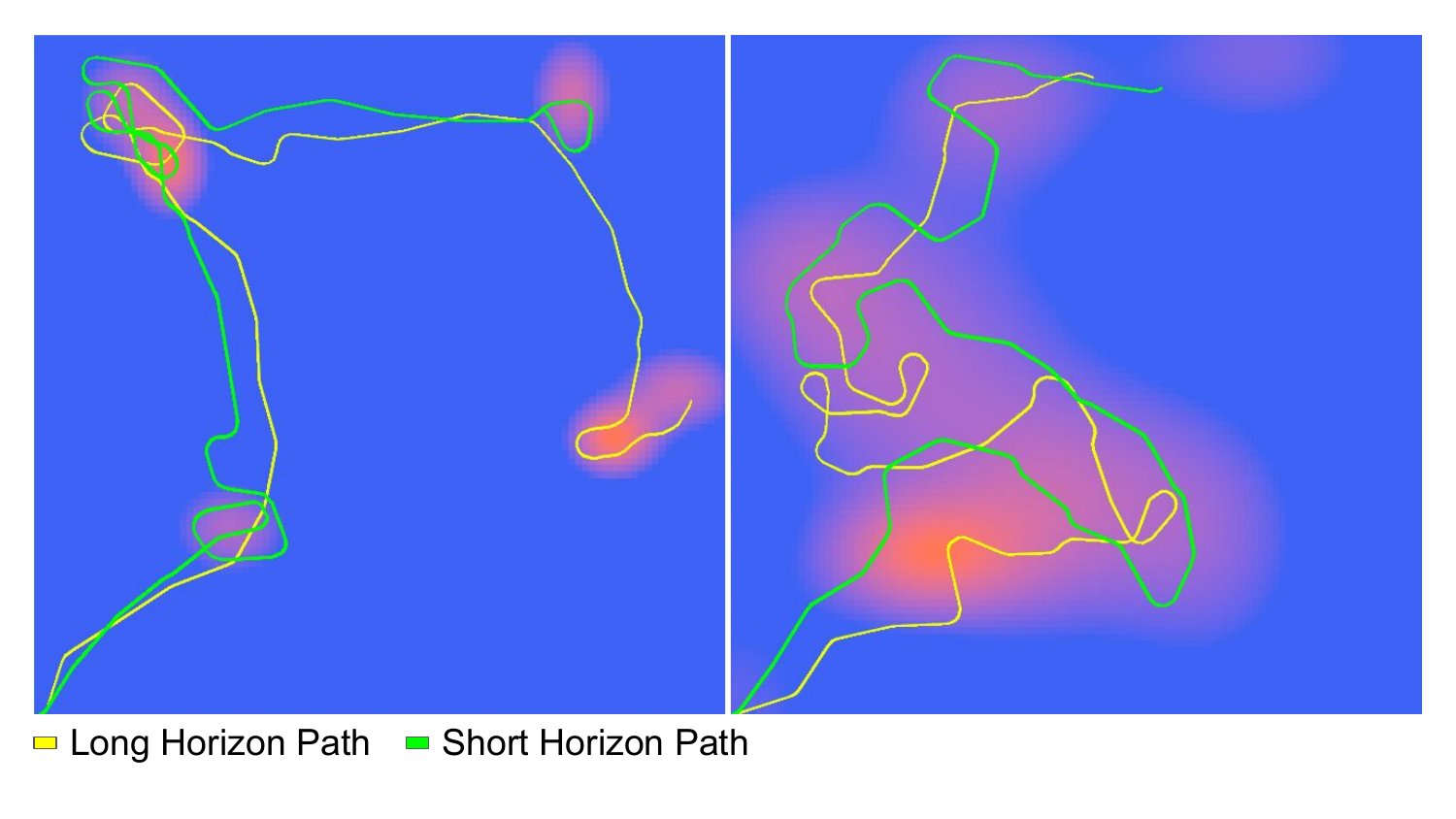}
    };
    \begin{scope}[x={(image.south east)},y={(image.north west)}]
        \fill[OrangeRed] (0.015, 0.083) -- +(3pt,0pt) arc (0:90:3pt) -- cycle;
        \fill[OrangeRed] (0.502, 0.083) -- +(3pt,0pt) arc (0:90:3pt) -- cycle;
    \end{scope}
\end{tikzpicture}
\caption{A comparison of IA-TIGRIS when planning over a long horizon versus a short horizon. Long-horizon planning enables IA-TIGRIS to allocate the entirety of its budget efficiently and is able to view all the high-entropy clusters in the example on the left. In the example on the right, short-horizon planning leads to less smooth paths with slightly higher information gain, as the high-entropy areas are connected and have lower entropy the farther they are from the starting position. }
\label{fig:horizon-analysis}
\end{figure}

To demonstrate the effects of long- and short-horizon planning, we ran a simulation on two different maps with varying information distributions as seen in Fig.~\ref{fig:horizon-analysis}, where the pink regions have high entropy and the blue regions have low entropy. The environment on the left side of the figure shows a distribution with clusters of information scattered across the space, while the environment on the right shows a denser distribution where the various areas of high entropy overlap. 
For short-horizon planning, \mbox{IA-TIGRIS} considers a planning horizon of up to $5000$ m, rather than planning over the entirety of the remaining budget. The long-horizon planning starts with the entire budget of 
$15000$ m and uses the entire remaining budget for every new plan in the planning process.

For the clustered belief map on the left, the long-horizon path searches over more clusters and achieves a higher overall information gain compared to the short-horizon path. This is because the short-horizon planner takes myopic actions, consuming the majority of its budget on the first two clusters. By the time it reaches the third cluster, it exhausts its budget and is unable to reach the last cluster. In contrast, the long horizon planner strategically balances its budget across all four clusters, enabling higher information gain.

For the dense belief map on the right, the long-horizon planner generates a smoother path and extends farther into the space than the short-horizon planner does. However, the long-horizon planner allocates a smaller portion of its budget on the highest information gain area near the start position. As a result, the short-horizon planner achieves a slightly better total information gain by prioritizing immediate rewards and concentrating its budget on the highest information cluster close to the starting position.  

From these two maps, we observe that the long-horizon planner excels in environments with clustered information where reasoning over long horizons is essential to avoid myopic choices and balance the allocation of budget. The short-horizon planner performs well in environments with densely distributed information, where following local gradients of information gain can lead to effective solutions. These findings suggest that the impact of the planning horizon on system performance depend on the environment. In our testing framework and applications, planning over the entire budget typically yields the best results, as our prior belief maps often feature clustered regions of high information gain.

%% file: tex/5_results.tex
\subsection{Baseline Planners}

To evaluate the performance of \PlannerName, we compare it against several baseline methods. In the following section, we describe these baselines, their implementation details, and their respective advantages and limitations, particularly in the context of information gathering in large, high-dimensional search spaces. The simulation framework and vehicle parameters remain consistent across all planners, and each method is allowed to replan during testing.

\subsubsection{MCTS}

MCTS can be a powerful technique for finding feasible and optimal paths in complex environments. It is a heuristic search algorithm that builds a search tree incrementally through repeated simulations. At each iteration, it selects a node to explore based on a selection policy (often the Upper Confidence Bound or UCB1 algorithm), expands the tree by adding possible actions from that node, runs a simulation from the newly added node, and updates the statistics of nodes along the path traversed during the simulation. 

The UCB1 algorithm is a technique commonly used in the context of multi-armed bandit problems and MCTS for balancing exploration and exploitation. This algorithm helps in selecting actions or nodes that are likely to yield high rewards while also exploring less-frequented options to gather more information about their potential rewards. 

We formulate our UCB score in the following manner, \\
\begin{equation*}
    UCB_\text{node} = \frac{I(X_{\text{node}})}{\alpha} + C \times \sqrt{\frac{\ln(N_\text{tree})}{N_\text{node}}}
\end{equation*}
Here $I(X_{\text{node}})$ denotes the estimated information gain from the node, $\alpha$ denotes the normalization factor which is given by $\frac{B}{v_\text{desired}}$, $B$ being the maximum planning budget and $v_\text{desired}$ being the desired speed of our UAV. The scalar $\alpha$ serves two complementary purposes: it aligns the units of the exploitation term with the exploration bonus, and it embeds mission‐level context—how much “time” the robot effectively has to collect information—directly into the scoring rule. Otherwise, the magnitude of $I(X_{\text{node}})$ can easily dominate the square–root term, leading the search to behave almost greedily. Intuitively, $\frac{B}{v_{\text{desired}}}$ is the maximum flight duration available for the plan; normalizing by this horizon rewards nodes that deliver high information density rather than simply high absolute information. When the operator increases the mission budget $B$, $\alpha$ grows and the exploitation term is damped, encouraging the planner to spend some of the extra allowance on exploration. Conversely, a higher desired speed $v_{\text{desired}}$ (which effectively shortens the decision horizon for a fixed $B$) decreases $\alpha$, sharpening the focus on immediate, high-value observations. Although we did not run a dedicated ablation to isolate the impact of normalization, qualitative tests during early integration were instructive: an unnormalized exploitation term consistently overpowered the exploration bonus, while introducing the 
$1/\alpha$ factor restored a healthy exploration–exploitation balance under every experiment setting we later report. We leave a more systematic study of normalization effects to future work. $C$ denotes the exploration weight, and $N_\text{tree}$ denotes the number of visits to the tree root node while $N_\text{node}$ denotes the number of times the present node has been visited.

\begin{figure}[t]
\centering
\includegraphics[trim={.7cm 0cm .5cm 1.4cm},clip,width=\columnwidth]{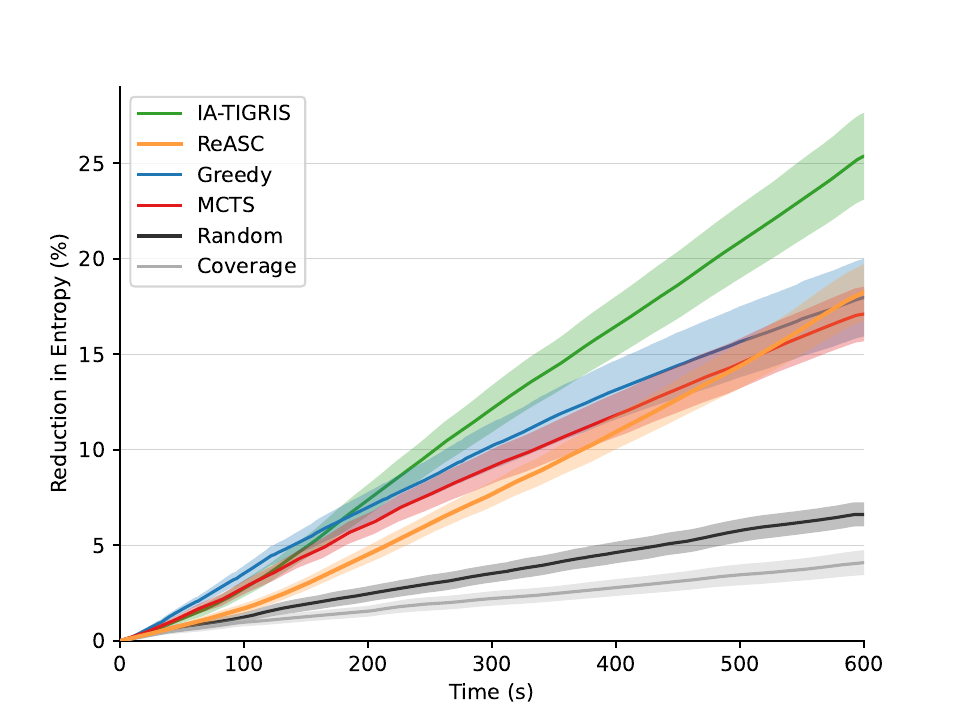}
\caption{The Monte Carlo simulation results for the planners. The plots show the average percent reduction in entropy over the course of the simulations, and the shading shows the 95\% confidence intervals. IA-TIGRIS outperforms all the baselines.}
\label{fig:mc_results}
\end{figure}

After selecting a candidate node, if it has been visited before, it is expanded by applying motion primitives to generate child nodes, growing the tree. Unvisited nodes skip this step. Following expansion, either the unvisited candidate node or one of its children is selected for the simulation phase, where the future values of nodes along the path are estimated to update the total potential information gain. The potential information gain informs the selection policy in subsequent iterations. Once planning time expires, the path with the highest information gain is returned.

While MCTS is probabilistically guaranteed to converge to the optimal path \cite{mcts_ref_1}, it is constrained to actions within a predefined set of motion primitives. Its reliance on random sampling to estimate the future value of nodes can result in poor approximations, particularly in environments with sparse, localized pockets of high information gain. This limitation is especially pronounced in large search areas or scenarios with large budgets constraints, where estimating future node values becomes increasingly expensive. As a result, in such scenarios, MCTS is often implemented with a finite planning horizon, which can restrict its ability to account for long-term consequences or dependencies in the environment.

\subsubsection{ReASC}

The Rapidly-exploring Adaptive Search and Classification (ReASC) algorithm \cite{hollinger_long-horizon_2015} is a sampling-based IPP method that extends the RIG-Tree algorithm \cite{hollinger_sampling-based_2014} with online replanning. It was originally demonstrated on an autonomous aquatic surface vehicle tasked with classifying regions that lie within a target depth using a grid-based probability map. In ReASC, samples are drawn uniformly in the continuous state space, and information gain is evaluated only at tree nodes, without accounting for rewards accumulated along edges.

In our setting, the belief map spans a much larger environment and each camera observation updates many cells along the continuous trajectory, in contrast with the ReASC test scenario where each measurement updates a single grid cell. To make ReASC tractable in our larger, longer-horizon tests, we incorporate the same belief-map node embedding used in \mbox{IA-TIGRIS} (Section~\ref{sec:embedding}), storing per-node belief deltas in a hash map while otherwise preserving the algorithm’s original sampling and reward structure. For replanning, ReASC is integrated into our common replanning framework, which is robust to variable segment lengths and planning times.

\subsubsection{Greedy}

For the greedy planner, we iterated through each cell within the search bounds and calculated the reward for a given cell $i$ as $g_i = R(X_i) / d_i$ where $R(X_i)$ is given through \eqref{equ:reward} and $d_i$ represents the Euclidean distance between the current position of the robot at the current time $t$ and the closest viewpoint to the cell. To compute this viewpoint, the yaw between the current pose of the robot and the intersected cell is first calculated. Using this yaw angle, the robot's sensor configuration, and  the desired flight altitude, we calculate the $x$ and $y$ coordinates that results in viewing the cell. With this formulation, the planner prioritizes regions with a high ratio of entropy to distance. This prioritization can lead to locally optimal choices that contradict with paths that lead to higher information gain over the entire trajectory.

\begin{figure*}[t]
    \centering
    \begin{subfigure}[b]{0.99\textwidth}
        \centering
        \includegraphics[trim={0cm 0.3cm 0cm 0cm},clip,width=\textwidth]{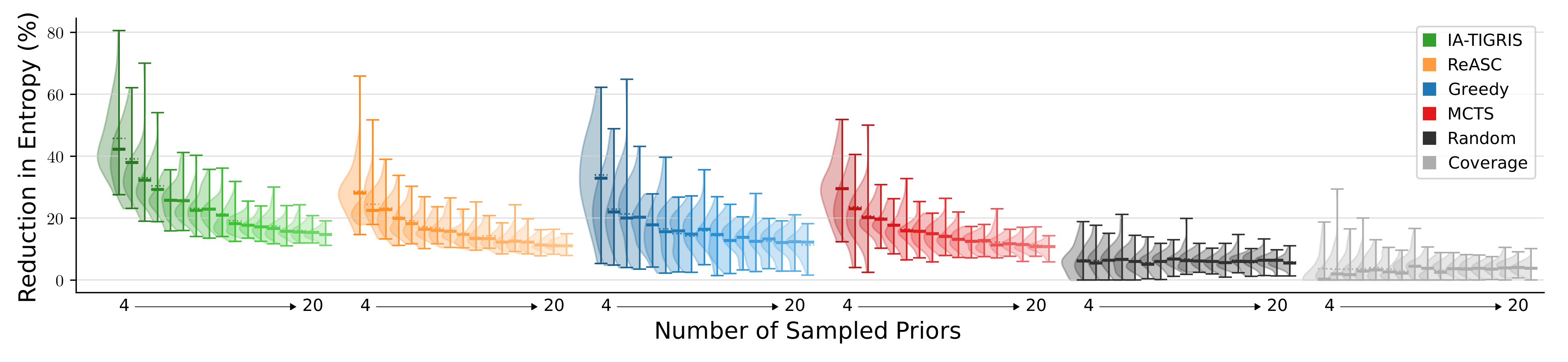}
    \end{subfigure}
    
    \begin{subfigure}[b]{0.99\textwidth}
        \centering
        \includegraphics[trim={0cm 0cm 0cm 0cm},clip,width=\textwidth]{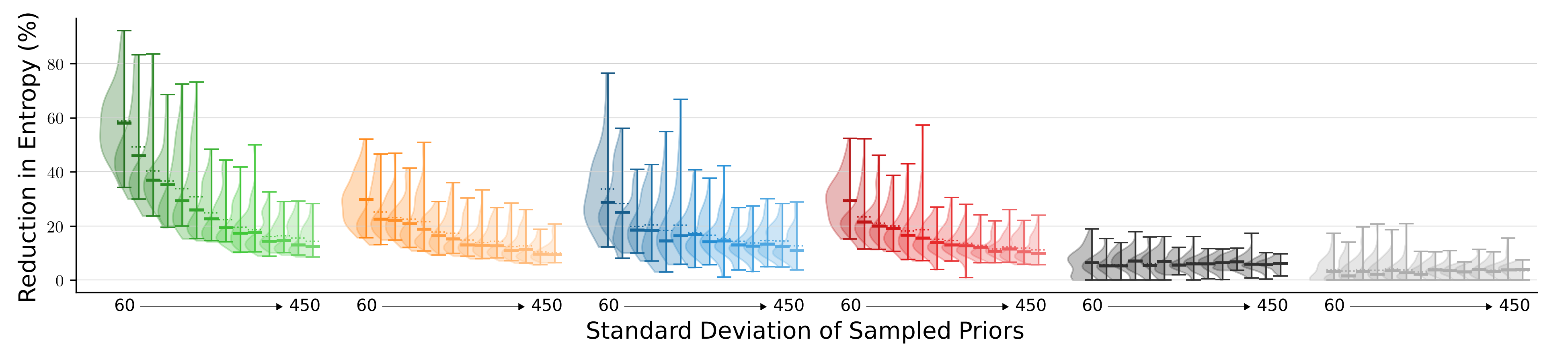}
    \end{subfigure}
    \caption{A comparison of the methods based on the number and standard deviation of the sampled prior clusters. The violin plot shows the distribution, with solid and dotted lines marking the median and mean. IA-TIGRIS is most effective compared to the baselines when there is high variation in the search space. As the search space prior information becomes more evenly spread out or uniform, the performance gap between the methods tends to decrease.}
    \label{fig:targets_sigmas}
\end{figure*}

\subsubsection{Random}

The random planner operates by iteratively sampling points within the defined search bounds and calculating the minimum-cost path to observe each sampled point. This process is repeated until the available budget is fully expended. The random planner does not utilize any prior information about the environment or target distribution. Additionally, it does not optimize the sequence of actions, instead treating each sampled point independently without considering the global structure of the search problem. This simplicity allows the random planner to highlight the performance benefits of more sophisticated methods by providing a lower-bound comparison for evaluation.

\subsubsection{Coverage}

The coverage planner generates a plan that systematically covers the entire search space using a straightforward lawn-mower pattern. The spacing between each pass is set to match the width of the projected observation footprint at 20\% from the bottom, ensuring that no grid cells are missed. This spacing also maintains a distance that enables high-quality sensor measurements. However, due to the size of the search spaces considered, the coverage planner spends significant time surveying empty regions. This approach results in inefficient use of the budget, as it prioritizes full coverage with safe sensor overlap, even in areas with little or no valuable information. While simple and robust, this method highlights the tradeoff between exhaustive coverage and efficient, targeted exploration.

\subsection{Tests and Analysis}
To evaluate the efficacy of IA-TIGRIS against baseline methods, we perform Monte Carlo testing and analyze the impact of the prior and budget on the performance of each method. In all test cases, rewards are calculated using \eqref{equ:reward}, and horizon lengths are set to match the full budget. The tests are conducted on an Intel Xeon CPU E5-2620 v4 @ 2.10GHz, ensuring consistent computational conditions across all evaluations.

\subsubsection{Monte Carlo Testing}
Our simulated testing environment is a $5000\times5000$ m square with Gaussian-distributed prior information randomly placed throughout the search space. 
The number of prior clusters was sampled uniformly from the interval $[4, 20]$, with standard deviations sampled from $[60, 450]~\text{m}$, and maximum values from $[0.05, 0.5]$.

The results of $100$ Monte Carlo tests are shown in Fig.~\ref{fig:mc_results}. IA-TIGRIS clearly outperforms the other methods, achieving a $38\%$ greater reduction in entropy than the next-best method. Early in the simulation, the greedy method initially gains information more quickly, as expected, but this result does not translate to better long-term performance because it optimizes only for immediate reward rather than total information gain over the full budget. ReASC is the next best-performing method, with results similar to greedy. Its performance is limited by its uniform sampling strategy and by evaluating information gain only at node locations. ReASC tends to spend substantial budget exploring low-value regions and fails to consider the information along the continuous trajectory. This stands in contrast to IA-TIGRIS, which focuses sampling on informative areas and leverages edge rewards to evaluate information collected between nodes, producing higher-quality long-horizon plans. 

MCTS performs slightly worse than the greedy approach, likely because it must build a tree of feasible sequences, while the greedy approach can select the next action in the entire belief map, even when it may be far away. This advantage of the greedy approach is particularly pronounced in sparse environments where MCTS may not explore distant, high-reward branches within its computational budget.

The random paths slightly outperformed the coverage paths. This result is likely because the lawnmower strategy requires sufficient overlap between passes to avoid missing areas, and its long straight paths often lead to redundant observations due to the UAV’s forward-facing camera. Changing the heading of the UAV is beneficial to viewing more of the search space, which may explain why random paths performed better.

We also conducted Monte Carlo tests where either the number of prior clusters or their standard deviation was held constant to analyze how variations in the belief map affect planner performance. The results, shown in Fig.~\ref{fig:targets_sigmas}, include two cases: the upper figure fixes the number of priors, while the lower figure fixes their standard deviation. All other agent and simulation parameters remained unchanged.

Across these tests, the performance gap between IA-TIGRIS and the baselines widens as the number and standard deviation of the Gaussian priors decrease. When entropy is more uniformly distributed across the search space, simpler methods perform reasonably well within the given budget. However, when information is concentrated in sparse, distinct regions, longer-horizon planning becomes essential. In such cases, \mbox{IA-TIGRIS} demonstrates a significant advantage by effectively reasoning about the budget and prioritizing high-value regions.

\subsubsection{Budget Analysis}
To evaluate the impact of budget constraints on performance, we conducted additional tests beyond our initial Monte Carlo experiments, evaluating budgets of $5000$ m, $10000$ m, $30000$ m, and $60000$ m. In this analysis, we also include a TIGRIS ablation, which generates only a single initial plan with no subsequent replanning, and we augment it with the same hash map belief embeddings used in \mbox{IA-TIGRIS} and ReASC. This variant isolates the contribution of adaptive, online refinement in \mbox{IA-TIGRIS} while ensuring a fair computational comparison. Table~\ref{tab:budgets} summarizes the average entropy reduction across these budgets.

\definecolor{tabfirst}{rgb}{0.73, 0.89, 0.73} %
\definecolor{tabsecond}{rgb}{0.83, 0.94, 0.83} %
\definecolor{tabthird}{rgb}{0.93, 0.98, 0.93} %

\begin{table}[t]
    \centering
    \resizebox{\linewidth}{!}{
    \begin{tabular}{l|ccccc}
    & $5000$ m & 10000 m  & 15000 m& 30000 m& 60000 m\\ \hline

    IA-TIGRIS  &  \cellcolor{tabfirst}$9.41\pm1.0$ &  \cellcolor{tabfirst}$18.28\pm1.8$ & \cellcolor{tabfirst}$25.36\pm2.3$ & \cellcolor{tabfirst}$41.08\pm2.9$ & \cellcolor{tabfirst}$58.85\pm2.9$ \\
    ReASC  &  \cellcolor{tabthird}$7.11\pm0.7$ & \cellcolor{tabthird}$13.38\pm1.2$ & \cellcolor{tabthird}$18.37\pm1.5$ & \cellcolor{tabsecond}$30.61\pm2.1$ & $45.25\pm2.4$ \\
    TIGRIS &  \cellcolor{tabsecond}$8.86\pm0.9$ & \cellcolor{tabsecond}$16.20\pm1.5$ & \cellcolor{tabsecond}$21.72\pm1.9$ & $27.80\pm2.2$ & $27.84\pm2.2$ \\
    Greedy  &  $6.99\pm0.8$ &  $13.10\pm1.5$ & $17.97\pm2.0$ & $30.00\pm2.3$ & \cellcolor{tabsecond}$49.38\pm3.5$ \\
    MCTS  &  
    $6.06\pm0.7$ &  
    $11.80\pm1.1$ & 
    $17.11\pm1.4$ & 
    \cellcolor{tabthird}$30.21\pm2.2$ & \cellcolor{tabthird}$48.68\pm2.7$ \\
    Random  &  $2.19\pm0.3$ & $4.29\pm0.7$ & $6.61\pm0.6$ & $17.50\pm1.2$ & $22.47\pm1.4$ \\
    Coverage  &  $1.58\pm0.3$ &  $2.82\pm0.4$ & $4.09\pm0.7$ & $12.04\pm1.9$ & $16.77\pm2.4$ \\

    \end{tabular}
    }
    \caption{Monte Carlo testing results given different budgets. The values are the average percent reduction in entropy and the 95\% confidence bounds. \mbox{IA-TIGRIS} had the best performance for all budgets.}
    \label{tab:budgets}
\end{table}

IA-TIGRIS consistently achieved the highest entropy reduction across all budget constraints, with a statistically significant margin over the other baseline methods. TIGRIS and ReASC formed the next tier of performance, with TIGRIS outperforming ReASC at lower budgets ($5000$--$15000$ m) and ReASC overtaking TIGRIS at higher budgets of $30000$ m and $60000$ m. Greedy and MCTS showed similar trends, with MCTS slightly outperforming Greedy at the $30000$ m budget, but Greedy achieving the second-highest entropy reduction at $60000$ m. At this highest budget, both Greedy and MCTS surpassed TIGRIS and ReASC. Consistent with our previous findings, the random and coverage planners yielded the lowest entropy reductions across all budgets.

The decline in TIGRIS performance at large budgets is likely due to the extremely long planning horizon, which prevents it from constructing a sufficiently long or high-quality plan within the limited planning time. This behavior underscores the importance of online refinement through adaptive planning, as demonstrated by IA-TIGRIS. ReASC also struggled at higher budgets, likely because the belief map becomes increasingly sparse as the map is observed over time. With uniformly random sampling and node-only reward evaluation, ReASC is less effective at locating the remaining high-information regions in the large, sparse search space.

Among the tested methods, IA-TIGRIS, ReASC, TIGRIS, and MCTS explicitly incorporate budget constraints into their planning algorithms. Notably, at lower budgets ($5000$ m and $10000$ m), these methods achieved higher entropy reduction compared to the equivalent time steps ($200$ s and $400$ s) in the $15000$ m budget scenario shown in Fig.~\ref{fig:mc_results}. This improvement reflects their ability to optimize total path reward under a fixed budget, in contrast to myopic next-best-action approaches such as the greedy method. The remaining methods---greedy, random, and coverage---maintain largely consistent behavior across budgets, as their planning strategies do not explicitly account for resource limitations.

The performance gap between IA-TIGRIS and the next-best baseline method (excluding the TIGRIS ablation) varied with budget size, showing margins of $32.3\%$, $36.6\%$, $38.1\%$, $34.2\%$, and $19.2\%$ in ascending budget order. This gap widened through the first three budget levels as problem complexity increased, before declining significantly at higher budgets. This performance pattern suggests that implementing a planning horizon could enhance efficiency by limiting tree search depth, enabling the planner to prioritize path quality optimization over exhaustive space exploration.

\subsection{Comparison Against Learning-Based Approach}\label{sec:learning}

We additionally compare IA-TIGRIS against a learning-based approach modified from CAtNIPP \cite{cao2023catnipp}.
CAtNIPP learns a policy to choose sampling locations from neighboring nodes in a prebuilt graph.
The original approach adopts a Gaussian-process-based information representation and defines information gain as the reduction of the traces of the covariance matrices.
We modified the algorithm to use our discrete grid-based representation, changed the reward to reflect information gain from the entire trajectory to the graph node, extended the state space from $(x,y)$ to $(x,y,\psi)$, and applied the identical camera parameters as IA-TIGRIS.

We trained the network using a $15000$ m budget on a $5000\times5000$ m map with randomly generated Gaussian distribution priors. Over $100$ Monte Carlo tests with a 200-node graph, the average uncertainty reduction was $15.78\!\pm\!1.9$.
This performance is $37.8\%$ worse than that of IA-TIGRIS, as presented in Table \ref{tab:budgets} column $3$, and slightly worse than MCTS and greedy.
A closer examination of the qualitative planning results highlights the limitations of the learning-based approach. Due to the sparsity of the graph, some high–information-gain regions are not represented by any nearby graph nodes, leading the policy to overlook these areas entirely. Moreover, even when distant regions contain substantially higher expected information gain, the learned attention-based policy exhibits a myopic bias: it consistently prefers moving to adjacent nodes with moderate gain rather than committing to multi-hop sequences required to reach more informative but distant regions. This suggests that the network fails to learn long-horizon trade-offs between short-term and future information gain, showing the difficulty of training effective policies when the sensing graph is either sparse or requires long-horizon reasoning.

Training with a larger graph of $1600$ nodes deteriorates the results to $9.85\!\pm\!1.2$ despite $48$ hours of training time, likely due to the inability of the network to learn an effective policy with a larger graph.
This deterioration reveals a fundamental trade-off that while larger graphs provide better spatial coverage, they become increasingly difficult to train effectively.
Therefore, the learning-based approach requires a carefully tuned graph size for specific problem settings and extensive training time to learn a good policy, which greatly limits its use cases.

\section{Field Deployments}\label{sec:field}

\subsection{Hexarotor Deployment}
The first field experiment that we present uses a hexarotor drone to cover an urban area shown in Fig.~\ref{fig:fig1}.
We designed this field experiment to simulate classifying where cars are within a search area.  
Hence, we set the plan request to focus on parking lots at the field test site (Fig.~\ref{fig:google_earth}, top), with the addition of three chosen grid cells within the parking lots being marked as having a higher uncertainty. The plan request boundaries and priors were created with GPS coordinates in Google Earth, exported as kml files and then converted into our plan request message format. 

The following sections details the hardware, autonomy, and experimental results for our hexarotor deployments.

\subsubsection{Hardware System}
The hardware consists of the DJI M600 Pro, shown in Fig.~\ref{fig:m600}, along with the physical sensing and onboard computer payload. The DJI M600 Pro contains a flight controller that handles pose estimation and position-based control. The DJI M600 Pro’s flight controller also handles teleoperation if human intervention is necessary. Beneath the drone's base, we mount a custom hardware payload.
That payload consists of an onboard computer, a Jetson Xavier, to run the autonomy software shown in Fig.~\ref{fig:functional_diagram}.
The payload also contains a downward-facing a camera for sensing the environment. The camera is a Seek S304SP thermal camera.
The camera intrinsics are used to calculate the frustum's intersection with the search map's cells in IA-TIGRIS.

\subsubsection{Autonomy System}
Fig.~\ref{fig:functional_diagram} illustrates the functional system diagram for the real-world field test on the DJI M600. The user specifies the initial plan request prior to takeoff. The \mbox{IA-TIGRIS} planner makes an initial plan on that plan request and sends a global path to the waypoint manager. The waypoint manager tracks the current waypoint within the plan and sends the next waypoint to the DJI software development kit, which then sends actuation commands to the motors. The position of the drone is used to calculate the distance from the drone to the ground and sends that distance parameter to the sensor model. The sensor model's true positive and false positive rate (derived from the detector testing data) is used to calculate the per-cell entropy updates in the search map manager. The search map manager publishes the current belief map, and the replanning node sends an updated plan request to the \mbox{IA-TIGRIS} planner every $10$ seconds.

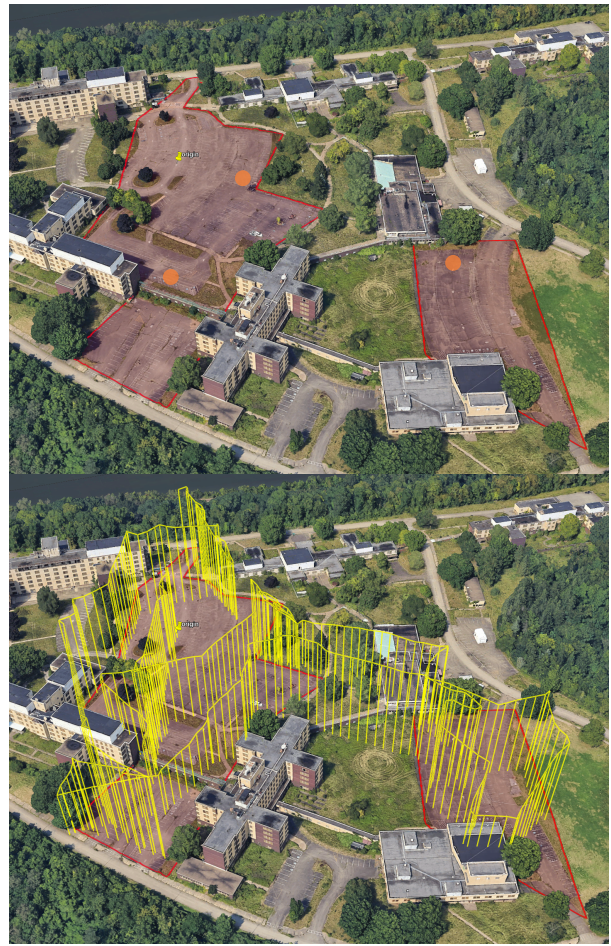
\begin{figure}[t] 
    \centering
    \renewcommand\arraystretch{0} %
    \setlength{\tabcolsep}{1pt} %
    \begin{tabular}{c}
        \begin{tikzpicture}
            \node[anchor=south west, inner sep=0] (image) at (0,0) {
                \includegraphics[width=0.9\linewidth]{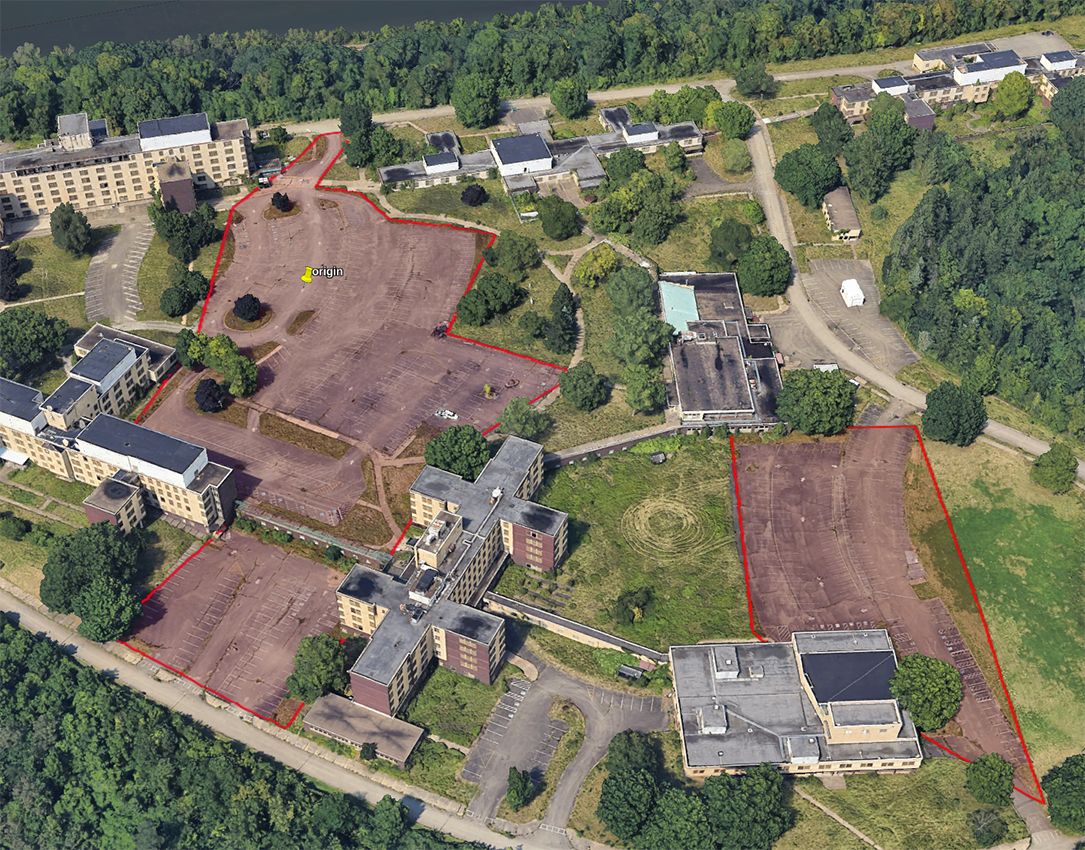}
            };
            \begin{scope}[x={(image.south east)},y={(image.north west)}]
                \fill[Orange, opacity=0.8] (0.74, 0.45) circle (3pt); %
                \fill[Orange, opacity=0.8] (0.27, 0.42) circle (3pt); %
                \fill[Orange, opacity=0.8] (0.39, 0.63) circle (3pt); %
            \end{scope}
        \end{tikzpicture} \\
        \\
        \includegraphics[width=0.9\linewidth]{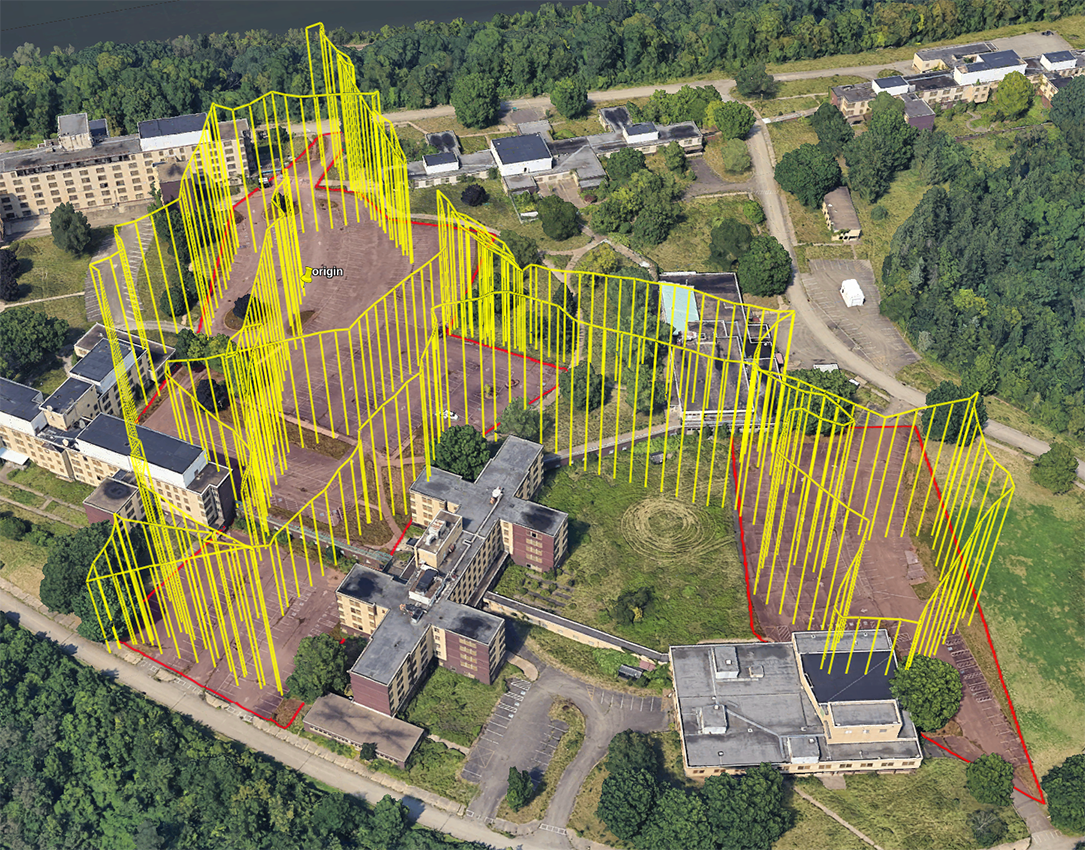} 
    \end{tabular}
    \caption{Google Earth screenshots illustrating the mission planning process and execution. Top: Areas of high entropy targeted for search are highlighted in red, representing regions with a binary occupied/unoccupied probability of 0.2. Three points of particular interest, each assigned a 0.5 probability, are marked in orange. Bottom: The executed drone flight path (yellow) shows the optimized path for maximum information gain across the search space.} 
    \label{fig:google_earth}
\end{figure}
\begin{figure}[t]
\centering
\includegraphics[width=\columnwidth]{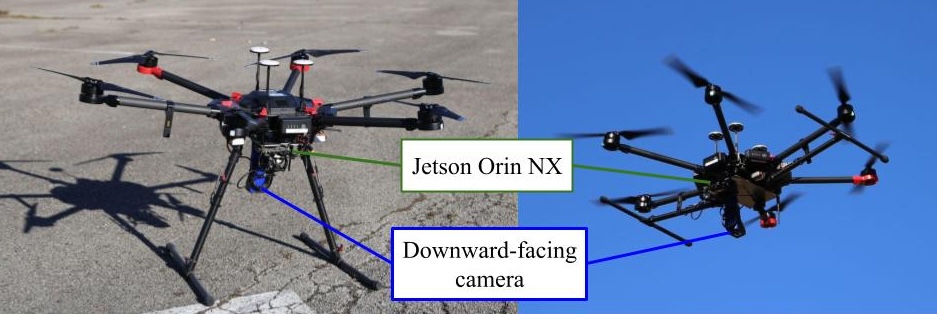}
\caption{Hexarotor system (DJI M600 Pro) with onboard computer and camera. Left image shows drone on the ground, right image shows drone in flight.}
\label{fig:m600}
\end{figure}
\begin{figure}[t]
\centering
\includegraphics[trim={0cm 0cm 0cm 0cm},clip,width=\columnwidth]{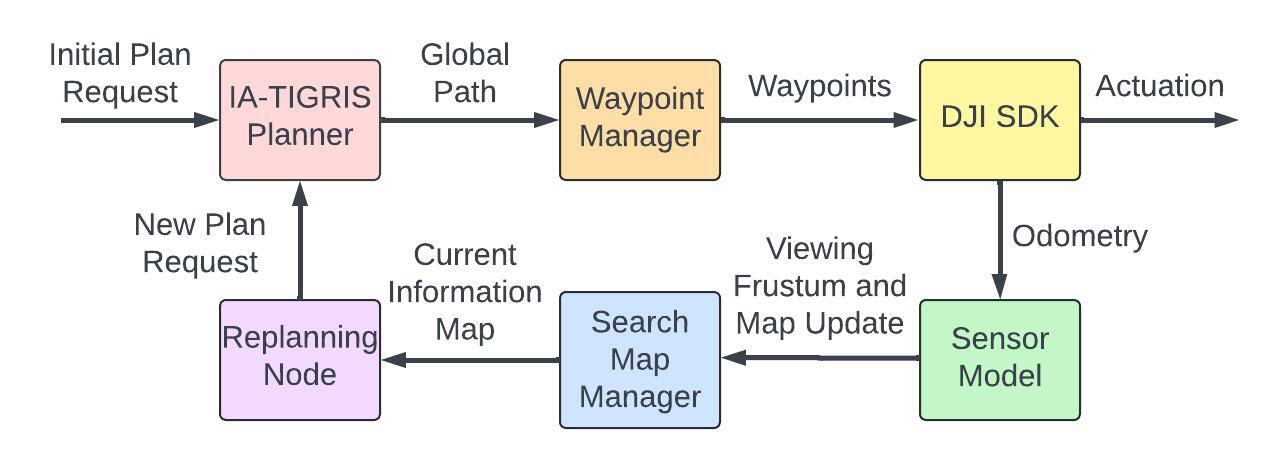}
\caption{Functional diagram of the DJI M600 Pro autonomy software.}
\label{fig:functional_diagram}
\end{figure}

The drone started at an altitude of $50$ m above the origin of the reference frame. The informed sampler in \mbox{IA-TIGRIS} was set to add states at altitudes of either $30$ m or $60$ m, creating a trade-off between observation area and detector accuracy. The budget was $2000$ m, the planning horizon was $600$ m, and the planning time was $10$ seconds.

\begin{figure}[t]
    \centering
    \begin{subfigure}[b]{0.48\columnwidth}
        \centering
        \includegraphics[width=1.0\linewidth]{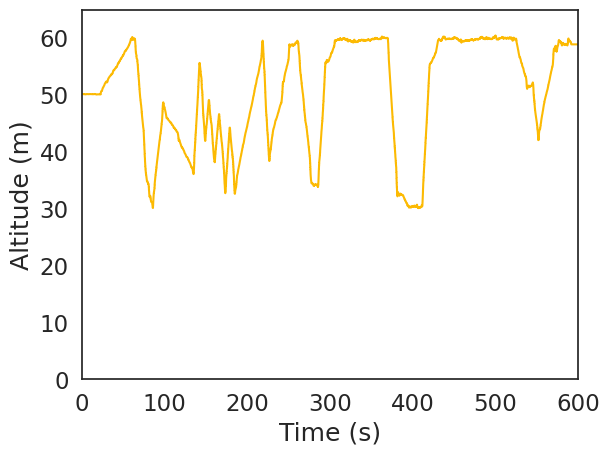}
        \caption{}
        \label{fig:m600_altitude_over_time}
    \end{subfigure}
    \begin{subfigure}[b]{0.48\columnwidth}
        \centering
        \includegraphics[width=1.0\linewidth]{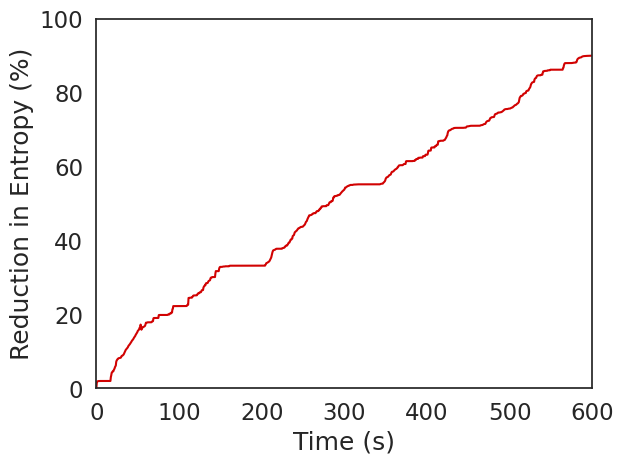}
        \caption{}
        \label{fig:m600_entropy_over_time}
    \end{subfigure}
    \caption{The results for our hexarotor field deployment. (a) Plot of flown altitude over time, showing large variation throughout the experiment. (b) Reduction in entropy percentage over time of field experiment.}
\end{figure}

\subsubsection{Experimental Results}

The bottom image of Fig.~\ref{fig:google_earth} shows the path selected by \mbox{IA-TIGRIS} in the search area. The figure highlights how the planner dynamically adjusts altitudes over time to balance coverage and sensing resolution, maximizing information gain. Higher altitudes allow for broader area coverage, while lower altitudes provide more detailed observations where needed. Additionally, the planner prioritizes revisiting the three regions of higher uncertainty, recognizing the need for repeated observations reduce entropy. This adaptive strategy ensures that uncertain areas receive sufficient attention to improve the belief map. As a result, the entropy of the belief map decreases to near zero by the end of the mission, as shown in Fig.~\ref{fig:m600_entropy_over_time}, indicating that the planner has effectively gathered the necessary information. This behavior demonstrates the planner’s ability to optimize sensing actions, balancing altitude selection, revisit frequency, and exploration to maximize mission success.

\begin{figure}[t]
\centering
\includegraphics[trim={4cm 4cm 0cm 4cm},clip,width=\columnwidth]{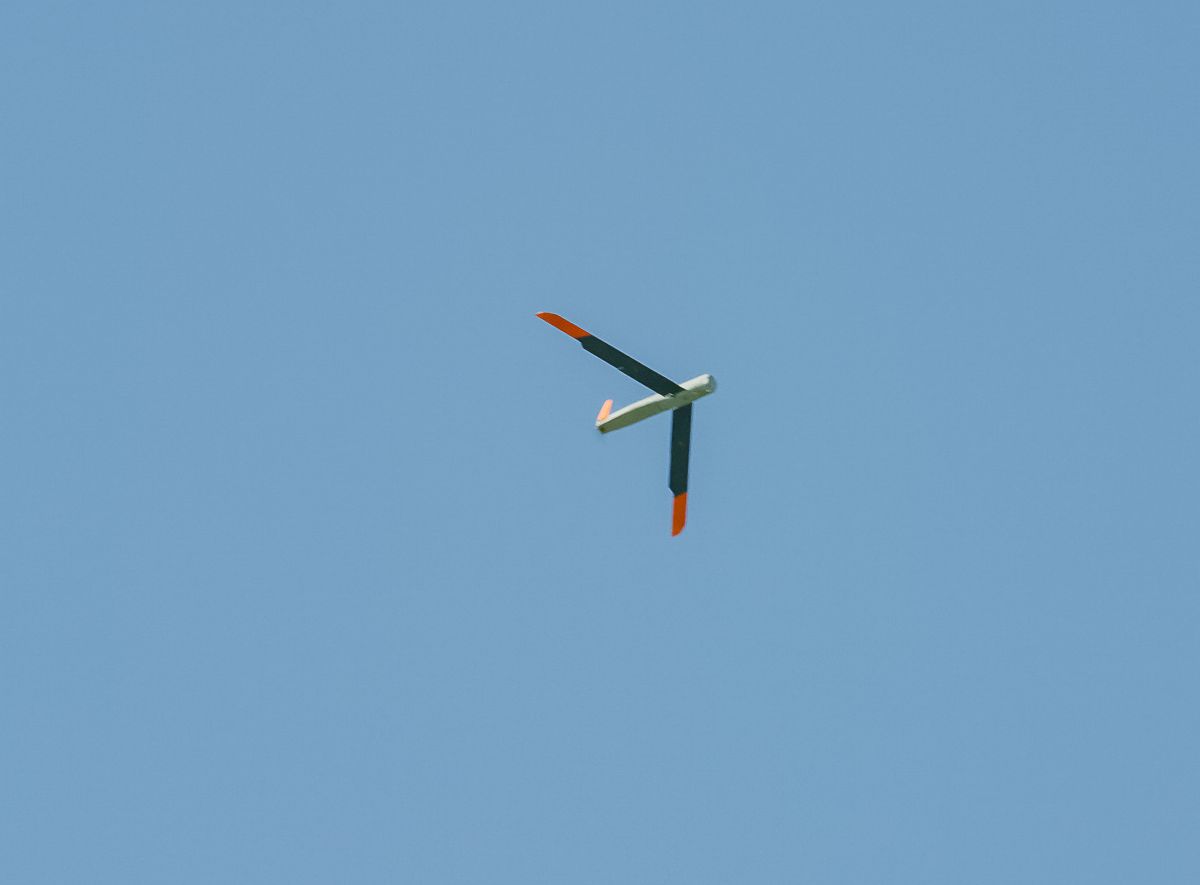}
\caption{Fixed-wing platform used for autonomous flights with an onboard camera pitched at 10 degrees\cite{alarewebsite}.}
\label{fig:tl1}
\end{figure}

\subsection{Fixed-Wing Deployments}

Our proposed approach was extensively tested on the fixed-wing AlareTech TL-1 UAV, shown in Fig.~\ref{fig:tl1}. The UAV is equipped with an onboard camera pitched at 10 degrees, which introduces a more challenging planning problem due to the nonholonomic motion model and the camera's field of view. Over more than 20 flight hours and 100 flights running IA-TIGRIS, we validated our approach with the objective to search for objects of interest in a large search space across a variety of test scenarios, including different terrain types, varying environmental conditions, and diverse target distributions. The object detector was trained on both real and synthetic data, while the sensor model was derived purely from synthetic data in order to get a detailed sensor model for a diverse set of range values and environmental conditions.

\begin{figure}[t]
\centering
\includegraphics[trim={3.0cm, 1.0cm, 3.0cm, 1.0cm},clip,width=\columnwidth]{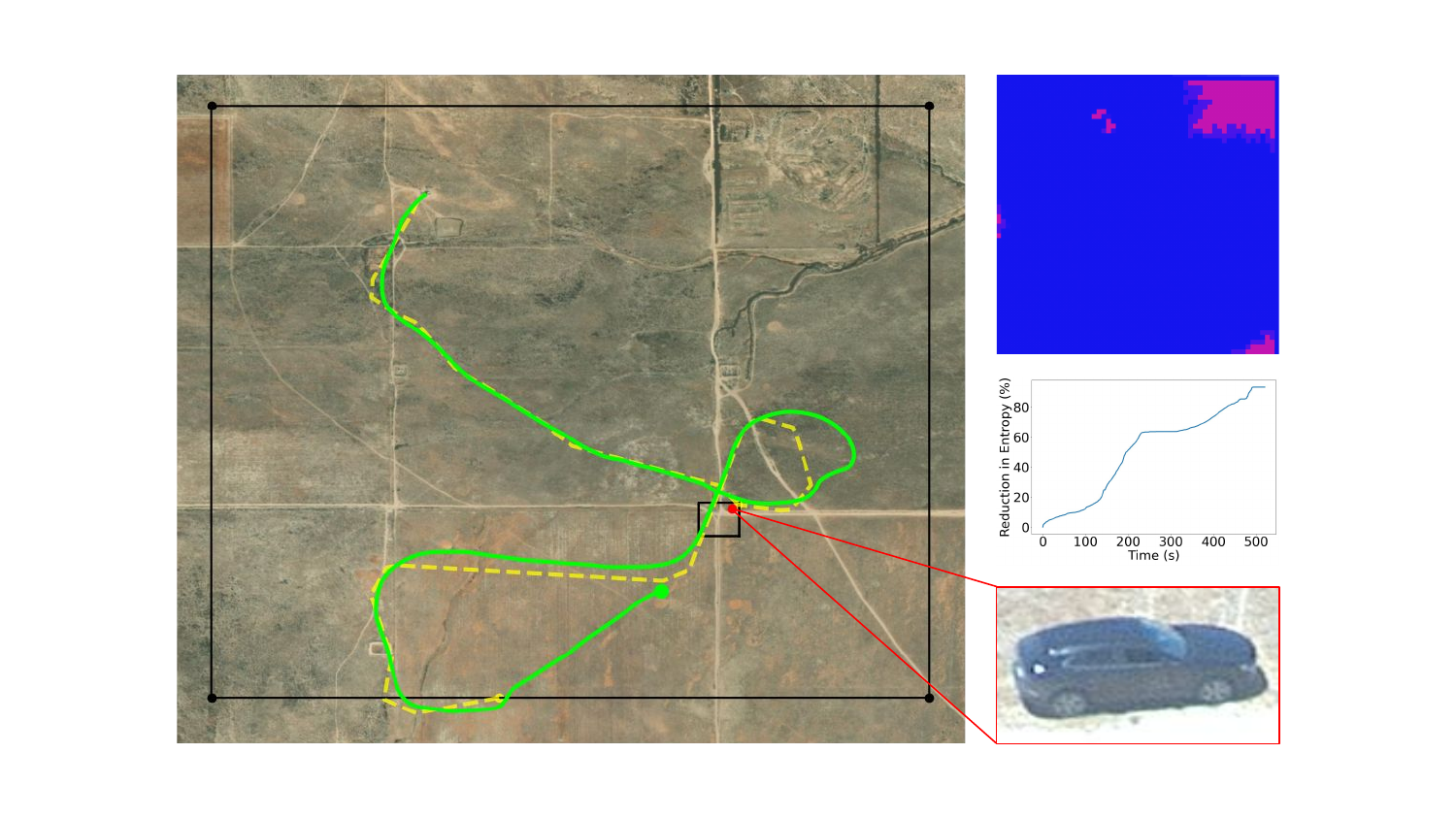}
\caption{An example path generated for the fixed-wing platform conducting a large-area search for an object of interest. The larger black rectangle denotes the search bounds, while the smaller black rectangle highlights a region of higher uncertainty. The red dot marks the estimated position of the detected object based on image detections. The upper-right map displays the information state after planning is complete, while the middle plot shows the percent change in entropy over mission time. The flown path illustrates a balance between allocating resources to the high-priority region and exploring other areas within the search space.}
\label{fig:fwd}
\end{figure}

An example mission from these tests is shown in Fig.~\ref{fig:fwd}. In this scenario, the planner was given the search bounds and a designated high-priority region. The resulting flight path prioritized revisiting the high-priority area twice, optimizing sensor use and ensuring maximum information gain. This strategy led to the successful detection of the object of interest, with its estimated position marked by the red dot in the figure. 

The map on the upper right in Fig.~\ref{fig:fwd} shows the belief map after plan execution was complete. Due to the UAV's limited budget, the upper right and lower left corners of the map were not searched by the agent. The budget is instead utilized to search over the area of higher priority two times. Compared to the paths in Fig.~\ref{fig:google_earth}, the paths for the fixed wing are smoother and have a larger turning radius, demonstrating how \mbox{IA-TIGRIS} respects the motion constraints of the vehicle. We can also see the effect of wind on the path execution, where the flown path shown in green deviates from the planned path shown in yellow. This path deviation illustrates the importance of online planning in the cases where this deviation is large or would accumulate over the course of a longer mission and cause the expected observed area differ significantly from the actual observed area.

%% file: tex/6_conclusion.tex
\section{Conclusion}\label{sec:conclusion}
This paper presents a novel sampling-based IPP planner for autonomous robots that is both adaptive and incremental. Our planner is shown to be effective and computationally efficient, making it suitable for onboard deployment. We conduct a thorough evaluation of individual planner components, providing generalizable insights for other IPP frameworks. Extensive simulation testing against baseline methods highlights the planner’s advantages, and we further validate its effectiveness through demonstrations on two robotic platforms.

Future work includes extending this approach to multi-agent deployments for object search and search-and-rescue operations in semi-urban environments. Further improvements could involve batch sampling and parallel processing to enhance computational efficiency, automatic parameter and horizon selection based on belief map characteristics, and trajectory optimization for smoother paths. Additionally, expanding the framework to support planning with a gimballed camera would enable more flexible and targeted sensing capabilities.